\documentclass[lettersize,journal]{IEEEtran}
\usepackage[utf8]{inputenc}
\usepackage{amsmath,amsfonts}
\usepackage{algorithmic}
\usepackage{algorithm}
\usepackage{array}
\usepackage[caption=false,font=normalsize,labelfont=sf,textfont=sf]{subfig}
\usepackage{textcomp}
\usepackage{stfloats}
\usepackage{url}
\usepackage{verbatim}
\usepackage{graphicx}
\usepackage{cite}
\usepackage{amssymb}
\usepackage{hyperref}

\hypersetup{hidelinks}
\graphicspath{{fig/}}
\begin{document}

\title{F-LIVO: A Feature-Level Geometric Measurement Framework for LiDAR--Inertial--Visual Odometry}

\author{Xinye~Dai, 
        Yapeng~Wang, 
        Shujie~Zhou, 
        Zihao~Wang, 
        Shiqi~Zheng, 
        Jiayu~Hu, 
        Yusheng~Wang, 
        Weiwei~Song, 
        Weiming~Tang 

\thanks{The authors are with the GNSS Center, Wuhan University, Wuhan 430079, China.(e-mail: xinyed@whu.edu.cn; wangyp@whu.edu.cn; sjzhou2020@whu.edu.cn; wzihao@whu.edu.cn; shiqizheng@whu.edu.cn;  hujiayu@whu.edu.cn; yushengwhu@whu.edu.cn; sww@whu.edu.cn; tangwm@whu.edu.cn)}

\thanks{This work was supported by the National Key Research and Development Program of China under Grant 2023YFB3907301. (Corresponding author: Weiming Tang)}

}


\maketitle

\begin{abstract}
Measurement-level LiDAR--Inertial--Visual Odometry (LIVO) can achieve high accuracy, and its practical resource requirements depend on the adopted association and map-maintenance strategy. Persistently attaching dense photometric patches or appearance attributes to mapped LiDAR points can introduce additional storage, association, and update costs. This paper presents F-LIVO, a filter-based LIVO framework that couples a visual--inertial subsystem and a LiDAR--inertial subsystem through feature-level point-to-plane constraints between visual points and LiDAR planes. To preserve cross-modal coupling when visual triangulation is weak, we further recover visual feature depth from the LiDAR map through local plane fitting and ray--plane intersection. In the LiDAR update, the covariance of each point-to-plane residual is adaptively scaled according to local planar reliability, thereby reducing overconfident updates on poorly conditioned surfaces. Because the proposed method relies on sparse geometric features rather than persistent dense photometric map attributes, it retains a compact map representation. {Experiments on multiple public datasets show competitive, sequence-dependent accuracy. Across the evaluated degradation cases, F-LIVO maintains continuous estimation with comparatively low accumulated end-to-end drift. Under the evaluated resource configuration, it achieves the lowest mean and peak {central processing unit (CPU)} usage, the lowest peak resident set size (RSS), and the lowest RSS growth rate among the evaluated implementations, while R2LIVE retains a lower mean RSS.} The source code will be made publicly available upon acceptance to facilitate reproducible research.\footnotemark
\end{abstract}
\footnotetext{https://github.com/xinyeDai/F-LIVO}


\section{Introduction}
\IEEEPARstart{A}{ccurate} and robust ego-motion estimation is a fundamental requirement for autonomous navigation systems operating in complex real-world environments. However, solutions relying on a single sensing modality inevitably suffer from intrinsic limitations. Vision-based methods are sensitive to illumination variations, motion blur, and textureless scenes, while LiDAR-only approaches may encounter geometric degeneracy in environments dominated by repetitive structures or large planar surfaces \cite{orbslam, loam}. These complementary failure modes have motivated extensive research on multi-sensor fusion to improve the reliability of state estimation.

LiDAR--Inertial Odometry (LIO) has emerged as a dominant solution by tightly coupling LiDAR measurements with inertial sensing, providing robustness to illumination changes and accurate geometric motion estimation \cite{lio_sam, fastlio, fastlio2}. Nevertheless, even state-of-the-art LIO systems may suffer from reduced observability in scenes with weak geometric excitation, such as long corridors or open areas, resulting in accumulated drift over extended trajectories \cite{fastlio2, fastlivo2}.

To alleviate these limitations, visual sensing has been increasingly incorporated into LiDAR--inertial systems, giving rise to LiDAR--Inertial--Visual Odometry (LIVO) frameworks. By exploiting complementary camera, LiDAR, and {inertial measurement unit (IMU)} measurements, LIVO systems aim to improve observability and robustness when individual sensing modalities degrade.

Recent LIVO systems have demonstrated improved estimation accuracy through increasingly tight coupling across sensing modalities, including measurement-level fusion strategies that jointly optimize LiDAR and visual observations \cite{r3live, fastlivo, fastlivo2}. {Measurement-level fusion is not inherently computationally inefficient, as FAST-LIVO2 demonstrates efficient measurement-update scaling. However, the complete resource footprint depends strongly on the adopted association and map-maintenance strategy. Persistently attaching dense photometric patches or appearance attributes to mapped LiDAR points can introduce substantial storage, association, and update costs. F-LIVO targets this source of complexity by representing visual--LiDAR coupling through sparse geometric constraints on a shared map.}

Recent studies have investigated multisensor odometry under challenging sensing and environmental conditions, including semantic-enhanced LIO in tree-rich scenes, global navigation satellite system (GNSS)--inertial--LiDAR fusion under sensor degradation, and neural inertial aiding for LiDAR--inertial odometry \cite{se_lio_tim, gao2024robust, wang2023lidar_tim, inin_lio_tim}. These studies reflect continued interest in exploiting complementary sensing information to improve state estimation under modality-specific degradation.

These observations suggest that tighter coupling alone is not always the optimal solution. Instead, a practical LIVO system should balance three requirements: effective cross-modal information exchange, explicit handling of measurement reliability, and bounded computational and memory cost. For practical multisensor odometry, this trade-off is closely related to how cross-modal geometric constraints are constructed and how their uncertainty is modeled.

In this work, we propose F-LIVO, a feature-level LiDAR--Inertial--Visual Odometry framework that uses structured geometric primitives instead of dense photometric associations. The method combines LiDAR point-to-plane constraints for scan-to-map registration with cross-modal point-to-plane constraints between camera-derived {three-dimensional (3D)} points and LiDAR local planes, and preserves these cross-modal constraints under weak parallax by recovering visual feature depth from the LiDAR map. Accordingly, heterogeneous measurements are unified into a common 3D point-to-plane geometric form, while the dynamic point-to-plane noise model is developed from LiDAR-supported point-to-plane constraints whose local plane quality can be directly evaluated from point-cloud neighborhoods. {F-LIVO targets the narrower complexity of persistent cross-modal map maintenance by expressing visual--LiDAR coupling through sparse geometric constraints on a shared map, without attributing the resource behavior of any one baseline to measurement-level fusion as a category.}

{The methodological distinction of F-LIVO is not the coexistence of a {multi-state constraint Kalman filter (MSCKF)-style} visual update and an {iterative extended Kalman filter (IEKF)-style} LiDAR update. These are established components. The distinctive measurement is the direct cross-modal point-to-plane residual constructed between a camera-derived 3D feature point and a local LiDAR plane. Consequently, visual and LiDAR observations are connected through an explicit geometric residual, rather than interacting only through a shared state estimate.}

{The main contributions of this work are summarized as follows:}

{(1) We introduce a direct cross-modal geometric measurement model that associates camera-derived 3D feature points with local LiDAR planes and incorporates the resulting point-to-plane residuals into a shared error-state filter. When visual triangulation is weak and an informative local LiDAR plane is available, a conditional ray--plane depth-recovery mechanism preserves the construction of these cross-modal constraints.}

{(2) We develop an adaptive covariance model for LiDAR point-to-plane residuals based on local planar reliability. A fixed-versus-adaptive ablation demonstrates that the proposed weighting improves estimation stability under the evaluated geometric degradation conditions.}

{(3) We maintain visual and LiDAR geometric information in a compact shared map representation and validate the resulting system on multiple public datasets, showing competitive, sequence-dependent accuracy; continuous estimation with comparatively low accumulated end-to-end drift across the evaluated degradation cases; and, under the evaluated resource configuration, the lowest mean and peak CPU usage, the lowest peak resident set size (RSS), and the lowest RSS growth rate among the evaluated implementations, while R2LIVE retains a lower mean RSS.}

\section{RELATED WORK}
\subsection{{Structured Visual--Inertial Odometry (VIO)}}
Existing VIO systems~\cite{vinsmono, openvins, orbslam3} have demonstrated strong accuracy and robustness by tightly coupling visual measurements with inertial sensing. Most early VIO methods primarily rely on point-based visual features, which perform well in richly textured environments but may suffer from reduced observability in scenes dominated by structural regularities or weak textures.

To address this limitation, several studies have explored the integration of higher-level geometric primitives. Representative approaches such as PL-VIO~\cite{plvio} and PL-VINS~\cite{plvins} extend the VINS-Mono framework by jointly optimizing IMU residuals together with point and line reprojection errors within a sliding-window optimization framework. By introducing line features, these methods enhance geometric constraints in structured environments and generally achieve higher localization accuracy than point-only VIO systems. However, this improvement is accompanied by increased computational overhead, as line feature detection and tracking are considerably more expensive than point-based processing.

PL-VINS improves the line segment detection strategy used in PL-VIO, reducing the average line extraction time to approximately 20~ms and enabling real-time operation. Similarly, PLC-VIO~\cite{plcvio} employs the EDLines algorithm for line feature extraction and reports a comparable computational cost of around 20~ms per frame. These results indicate that carefully designed line detection pipelines can partially alleviate the computational burden while preserving the benefits of line-based constraints.

Beyond point and line features, planar primitives have also been incorporated into VIO systems to further exploit scene structure. In~\cite{mesh}, a non-iterative plane extraction and mesh generation method is proposed, allowing plane features to be extracted in less than 1~ms. Building upon this framework, PLP-VIO~\cite{plp_vio} augments PL-VINS by incorporating line information into the reconstructed mesh and introducing explicit line-to-plane constraints, resulting in improved estimation accuracy. In a different direction, PVI-DSO~\cite{pvidso} presents a direct sparse visual--inertial odometry approach that leverages planar regularities to improve robustness.

It is worth noting that many plane-based methods impose structural assumptions during plane extraction, typically restricting planes to horizontal or vertical orientations. To relax such assumptions, ov\_plane~\cite{ov_plane} proposes a point-based VIO framework with planar regularities under the MSCKF formulation. By avoiding the explicit inclusion of planes in the state vector, this approach incorporates planar constraints without incurring significant computational spikes as the number of constraints increases.

We use ov\_plane as the VIO subsystem and modify its Extended Kalman Filter (EKF) update by introducing LiDAR point clouds to construct additional point-to-plane constraints with the point and plane features in the VIO system.

\subsection{LiDAR-Inertial Odometry}
Early LiDAR-Inertial odometry systems evolved from feature-based LiDAR odometry frameworks represented by {LiDAR odometry and mapping (LOAM)}~\cite{loam}, including representative extensions such as LeGO-LOAM~\cite{lego_loam}. These methods estimate sensor motion by extracting edge and planar features from LiDAR point clouds and performing scan-to-scan and scan-to-map registration. By exploiting the local smoothness of individual scan lines, feature selection effectively reduces the computational burden of pose estimation, while inertial measurements are mainly used to provide motion prediction and compensate for motion distortion, improving robustness under rapid motion. Despite their effectiveness, such pipelines primarily rely on loosely coupled sensor fusion and local geometric optimization, which limits long-term global consistency.

As point-to-plane residuals became a dominant registration model, subsequent research began to explore more explicit representations of local surface geometry. Surfel-based mapping systems such as SuMa++~\cite{suma_pp} maintain collections of small planar surface elements and register incoming LiDAR scans by optimizing point-to-plane residuals against these surfels. While this representation improves geometric stability and registration accuracy, surfel association typically relies on projecting or rendering surfels onto virtual image planes, which introduces considerable computational overhead and limits scalability in large-scale or long-term mapping scenarios.

To overcome the limitations of locally optimized pipelines in terms of global consistency, optimization-based tightly coupled LiDAR--Inertial odometry methods were proposed. LIO-SAM~\cite{lio_sam} integrates LiDAR scan matching and IMU preintegration within a factor-graph-based smoothing and mapping framework, enabling loop closure and globally consistent state estimation in large-scale environments. This formulation also facilitates the fusion of additional sensors, such as GNSS. However, the batch optimization strategy incurs increased computational and memory costs, which may challenge real-time performance in long-duration applications.

In parallel with optimization-based approaches, filter-based tightly coupled LIO frameworks have been actively investigated to improve computational efficiency while maintaining high estimation accuracy. FAST-LIO~\cite{fastlio} and FAST-LIO2~\cite{fastlio2} adopt an iterated error-state Kalman filter, where the system state is propagated using high-rate inertial measurements and updated by directly incorporating LiDAR point-to-plane residuals. These methods avoid explicit feature extraction and perform scan-to-map registration against an incrementally maintained local map using efficient spatial data structures, enabling high-frequency state estimation with real-time performance.

Along another line of research, voxel map--based LiDAR and LiDAR--Inertial {simultaneous localization and mapping (SLAM)} methods discretize the environment into voxelized spatial cells and maintain local geometric representations within each voxel to support efficient point-to-plane registration. Representative systems such as VoxelMap++~\cite{voxelmap_pp} and Voxel-SLAM~\cite{voxelSLAM} demonstrate that voxelized mapping can provide an effective trade-off between computational efficiency and geometric fidelity in large-scale real-time estimation. Furthermore, recent studies explore adaptive and probabilistic voxel-based plane representations to further improve robustness and accuracy under varying scene structures~\cite{adaptive_voxel_plane}.

Recent TIM publications further extend LIO robustness in application-driven settings. SE-LIO~\cite{se_lio_tim} improves state estimation in tree-rich environments through semantic filtering and cylinder-based geometric modeling. Gao \emph{et al.}~\cite{gao2024robust} and Wang \emph{et al.}~\cite{wang2023lidar_tim} incorporate GNSS information to enhance robustness during sensor degradation and long-duration operation, while ININ-LIO~\cite{inin_lio_tim} explores neural inertial aiding for LiDAR--inertial odometry. These studies emphasize robustness enhancement within the LIO subsystem itself, but they do not explicitly address lightweight cross-modal feature coupling with visual plane constraints.

We use FAST-LIO2 as the LIO subsystem and refine its point-to-plane noise model so that the residual covariance better reflects the local geometric conditions.

\subsection{LiDAR--Inertial--Visual Odometry}
LiDAR-inertial-visual fusion systems can generally be categorized into two paradigms: loose coupling and tight coupling. Loose coupling approaches~\cite{zhang_singh, stereo_lidar} typically utilize LiDAR measurements to provide depth information for camera images, thereby producing an effect similar to that of {red--green--blue plus depth (RGB-D)} cameras. During state estimation, these methods usually employ the output of a visual--inertial odometry (VIO) subsystem as the initial guess for LiDAR scan matching. While such designs are relatively straightforward to implement, their performance is often constrained by the limited interaction between different sensing modalities.

In contrast, tightly coupled approaches establish cross-sensor associations at different levels of the estimation pipeline. LVI-SAM~\cite{lvisam} performs fusion within a tightly coupled smoothing and mapping framework based on factor graph optimization and introduces a switching mechanism that allows individual subsystems to operate independently when faults are detected. {LIC-Fusion and LIC-Fusion~2.0~\cite{licfusion, licfusion2} incorporate IMU measurements, visual point features, and LiDAR plane features within an MSCKF framework. Their visual and LiDAR residuals constrain a common state, but remain modality-specific observations; no direct residual is constructed by associating an individual visual feature with a local LiDAR plane. F-LIVO introduces this explicit visual-point-to-LiDAR-plane association as a cross-modal geometric measurement.}

{R2LIVE~\cite{r2live} further strengthens the coupling among LiDAR, IMU, and camera sensors by combining sparse visual reprojection residuals and LiDAR geometric residuals within an error-state iterative Kalman filter framework. These modality-specific residuals constrain a common state, but R2LIVE does not construct a direct point-to-plane residual between an individual visual feature and a local LiDAR plane.}

{R3LIVE~\cite{r3live} and FAST-LIVO2~\cite{fastlivo2} are representative measurement-level systems that establish direct visual--LiDAR coupling mainly through photometric associations between images and mapped LiDAR points carrying appearance information. F-LIVO instead constructs sparse geometric associations between visual feature points and LiDAR local planes. The distinction therefore lies in the cross-modal residual and the information maintained by the map, rather than simply in whether the system is tightly coupled. The resource cost of either paradigm depends on its implementation; measurement-level fusion itself is not inherently inefficient.}

{Closely related feature-level {LiDAR--visual--inertial odometry (LVIO)} approaches include HDA-LVIO~\cite{hda_lvio} and the tightly coupled monocular visual--LiDAR odometry and mapping method proposed by Shu and Luo~\cite{multimodal_feature_vlo}. HDA-LVIO extracts planes from a LiDAR map, projects their centroids into the image, estimates visual-feature depth over a sliding window, and forms hybrid associations through image-space reprojection constraints. Shu and Luo recover visual-feature depth and combine visual and LiDAR feature matching or registration in frame-to-frame odometry and multi-modal mapping. These studies establish prior use of LiDAR-supported visual depth and feature-level cross-modal association. F-LIVO is therefore distinguished not by these general ideas alone, but by its specific measurement representation: a shared 3D visual landmark is coupled directly to a local LiDAR plane through a point-to-plane residual within a unified error-state filter, using a sparse geometric map representation.}

Recent TIM works have also explored robustness-oriented LiDAR--visual--inertial designs from different perspectives. Dynam-LVIO~\cite{dynam_lvio_tim} improves LVIO performance in dynamic urban scenes through object-aware modeling, while the tightly coupled LIVO system in~\cite{liv_dual_tim} targets LiDAR with a limited {field of view (FoV)} by enhancing both LiDAR and visual subsystems. In addition, intermittent VIO-assisted LiDAR SLAM~\cite{vio_assisted_lidar_tim} mitigates LiDAR degeneracy by selectively invoking visual assistance. Although these methods improve robustness in specific scenarios, they do not construct the compact feature-level visual-point-to-LiDAR-plane constraints pursued in this work.

In this work, we do not adopt line or edge features for feature-level fusion. First, extracting line segments using visual methods typically incurs significant computational cost. Second, prior studies have shown that edge features contribute limited improvements in accuracy~\cite{ba_lidar}. Instead, by parameterizing structured features in the environment, our method jointly optimizes features using observations from both cameras and point clouds, thereby establishing direct constraints between the two sensors at the feature level. {This design avoids persistent dense photometric attributes on mapped LiDAR points while retaining sparse visual, LiDAR, and cross-modal geometric constraints within the unified estimator.}

\section{OVERVIEW OF THE PROPOSED SYSTEM}
\begin{figure*}[htbp!]
    \centering
    \includegraphics[width=\linewidth]{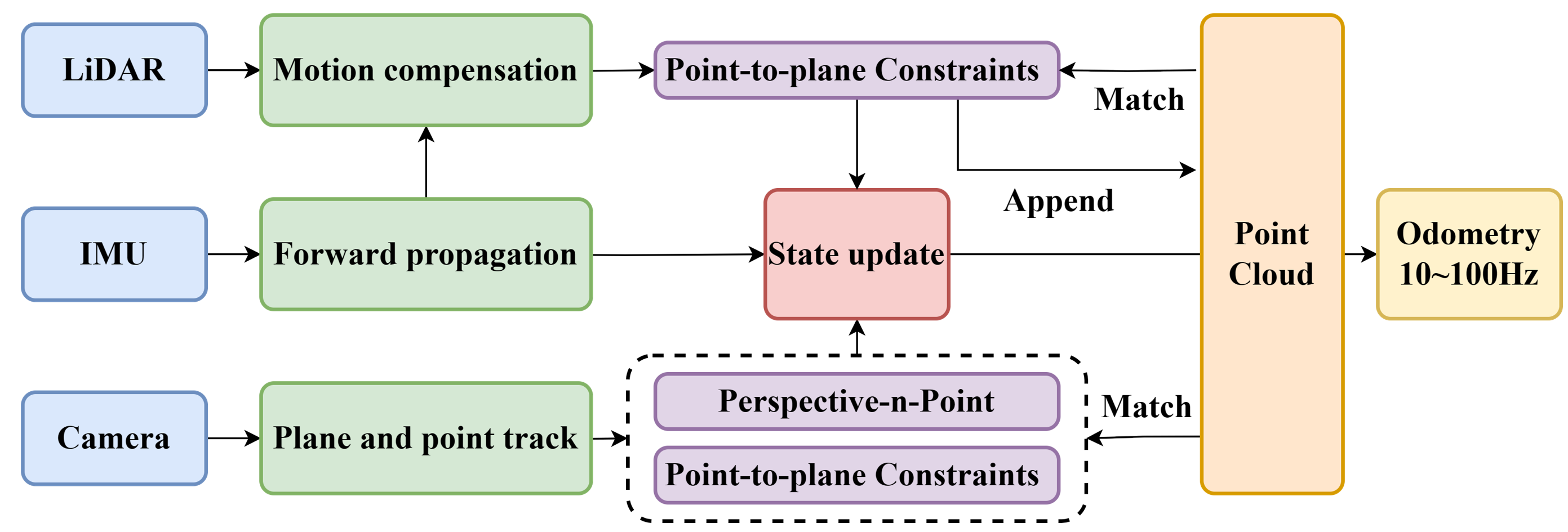}
    \caption{Overview of the proposed system. The system integrates LiDAR, IMU, and camera measurements within a unified filtering framework for LIVO. LiDAR and visual observations are coupled through point-cloud geometric information to achieve feature-level fusion. The resulting updates are applied to the relevant states in the filter, and the final output is the fused odometry estimate.}
    \label{fig:system_overview}
\end{figure*}

The proposed system tightly integrates LiDAR, camera, and IMU measurements within a unified filtering framework, as illustrated in Fig.~\ref{fig:system_overview}. During initialization, a short stationary interval is used to estimate the gravity direction and IMU biases and to initialize a gravity-aligned local odometry frame. The horizontal orientation of this frame is an initial yaw-gauge choice rather than an absolute-heading measurement. During normal operation, IMU measurements are first used to propagate the system state forward in time. LiDAR and visual observations are then incorporated to update the filter states through geometric and feature-level constraints. Specifically, the LIO subsystem uses undistorted point clouds to establish point-to-plane geometric constraints with the map, while the VIO subsystem adopts an MSCKF-based sliding-window formulation to utilize point and plane features. More importantly, cross-modal associations are established between visual features and point-cloud geometric structures, enabling feature-level fusion across sensing modalities. The resulting updates are applied to all relevant states in the filter, and the final output of the system is the fused odometry estimate.

{Sharing the state vector and covariance is necessary for unified filtering but is not, by itself, the proposed feature-level coupling. The coupling is established by the explicit point-to-plane residual between a camera-derived feature point and a LiDAR local plane.}

\subsection{Notation}
The main notations and symbol conventions used in this paper are summarized in Table~I, which serve as a reference for the formulations presented in the following sections.

\begin{table}[h]
\renewcommand{\arraystretch}{1.25}
\setlength{\tabcolsep}{6pt}
\caption{Important Notations}
\centering
\begin{tabular}{l p{0.72\columnwidth}}
\hline
\textbf{Notations} & \textbf{Explanation} \\
\hline
$\boxplus / \boxminus$ 
& The $\boxplus$ and $\boxminus$ operators on the state manifold \\

$(\cdot)^{G}$ 
& A vector $(\cdot)$ expressed in the global frame\\

$(\cdot)^{C}$ 
& A vector $(\cdot)$ expressed in the camera frame\\

$(\cdot)^{I}$ 
& A vector $(\cdot)$ expressed in the IMU frame\\

$(\cdot)^{L}$ 
& A vector $(\cdot)$ expressed in the LiDAR frame\\

$\mathbf{T}_{L}^{I}$ 
& Rigid extrinsic transformation from the LiDAR frame to the IMU frame \\

$\mathbf{T}_{C}^{I}$ 
& Rigid extrinsic transformation from the camera frame to the IMU frame \\

$\mathbf{x},\ \hat{\mathbf{x}}$
& The true and estimated states $\mathbf{x}$ \\

$\tilde{(\cdot)}$
& Linearized error-state quantity of $(\cdot)$ \\

$\breve{(\cdot)}$
& Compressed quantity after measurement compression \\

$\mathbf{z}$ 
& The measurement or residual vector \\

$\sigma_{\mathrm{plane}}$ 
& Local surface planarity score \\

\hline
\end{tabular}
\end{table}

\subsection{State vector}
The system state at time $t_k$ consists of the current IMU state $\mathbf{X}_{I_k}$, a set of IMU clones from past time instants $\mathbf{X}_{C}$, and the states of point features $\mathbf{X}_{f}$ and plane features $\mathbf{X}_{\pi}$ in the environment.

\begin{equation}
\label{eq:system_state}
\mathbf{x}_{k} =
\left[
\begin{array}{llll}
\mathbf{x}_{I_{k}}^{T} &
\mathbf{x}_{C}^{T} &
\mathbf{x}_{f}^{T} &
\mathbf{x}_{\pi}^{T}
\end{array}
\right]^{T}
\end{equation}

\begin{equation}
\label{eq:imu_state}
\mathbf{x}_{I_{k}} =
\left[
\begin{array}{lllll}
(\bar{q}_{G}^{I_{k}})^{T} &
(\mathbf{p}_{I_{k}}^{G})^{T} &
(\mathbf{v}_{I_{k}}^{G})^{T} &
\mathbf{b}_{g}^{T} &
\mathbf{b}_{a}^{T}
\end{array}
\right]^{T}
\end{equation}

\begin{equation}
\label{eq:imu_clones}
\mathbf{x}_{C} =
\left[
\begin{array}{lll}
\mathbf{x}_{T_{k}} & \cdots & \mathbf{x}_{T_{k-c}}
\end{array}
\right]^{T}
\end{equation}

\begin{equation}
\label{eq:point_feature_state}
\mathbf{x}_{f} =
\left[
\begin{array}{lll}
(\mathbf{p}_{f_{1}}^{G})^{T} & \cdots & (\mathbf{p}_{f_{i}}^{G})^{T}
\end{array}
\right]^{T}
\end{equation}

\begin{equation}
\label{eq:plane_feature_state}
\mathbf{x}_{\pi} =
\left[
\begin{array}{lll}
(\boldsymbol{\Pi}_{1}^{G})^{T} & \cdots & (\boldsymbol{\Pi}_{j}^{G})^{T}
\end{array}
\right]^{T}
\end{equation}

where $\bar{q}_{G}^{I_{k}}$ is the unit {Jet Propulsion Laboratory (JPL)} quaternion representing the rotation matrix from the global frame to the IMU frame at time instant $t_{k}$, $\mathbf{p}_{I_{k}}^{G}$ and $\mathbf{v}_{I_{k}}^{G}$ denote the position and velocity of IMU at time instant $t_{k}$, and $\mathbf{b}_{g}$ and $\mathbf{b}_{a}$ are the biases of the gyroscope and accelerometer. $\mathbf{x}_{T_{k}}=\left[\begin{array}{ll}
(\bar{q}_{G}^{I_{k}})^{T} & (\mathbf{p}_{I_{k}}^{G})^{T}
\end{array}\right]$. $\mathbf{p}_{f}^{G}$ represents the state of the SLAM points in the global frame. Plane feature $\boldsymbol{\Pi}^{G}$ is expressed as the minimal error-state closest point (CP)\cite{lips} in the global frame, which can be defined using the plane's normal vector $\mathbf{n}^{G}$ and distance scalar $d^{G}$ as $\boldsymbol{\Pi}^{G}=(\mathbf{n}^{G}, d^{G})$.

For state variables, we define their operations using "\(\boxplus\)" and "\(\boxminus\)". For a state $v$ in the vector space, these reduce to Euclidean addition and subtraction, while for quaternions they are given by the JPL quaternion operations\cite{trawny}.

In addition, the IMU, LiDAR, and camera are assumed to be rigidly mounted, and the extrinsic parameters $\mathbf{T}_{L}^{I}$ and $\mathbf{T}_{C}^{I}$ are assumed to be known. {The camera, LiDAR, and IMU streams are temporally synchronized before being processed by the proposed framework; residual temporal offsets are not estimated as part of the current state.}

\subsection{IMU Propagation and Initialization}
The raw accelerometer and gyroscope measurements, $\mathbf{a}_{m}^{I}$ and $\boldsymbol{\omega}_{m}^{I}$, are modeled as
\begin{equation}
\label{eq:acc_measurement}
\mathbf{a}_{m}^{I}
=
\mathbf{a}_{k}^{I}
+
\mathbf{R}_{G}^{I_{k}}\,\mathbf{g}^{G}
+
\mathbf{b}_{a_{k}}
+
\mathbf{n}_{a},
\end{equation}

\begin{equation}
\label{eq:gyro_measurement}
\boldsymbol{\omega}_{m}^{I}
=
\boldsymbol{\omega}_{k}^{I}
+
\mathbf{b}_{g_{k}}
+
\mathbf{n}_{g},
\end{equation}
where $\mathbf{a}_{k}^{I}$ and $\boldsymbol{\omega}_{k}^{I}$ denote the true specific force and angular velocity expressed in the IMU frame, $\mathbf{b}_{a_{k}}$ and $\mathbf{b}_{g_{k}}$ are the accelerometer and gyroscope biases, and $\mathbf{n}_{a}$ and $\mathbf{n}_{g}$ are zero-mean white Gaussian noises with $\mathbf{n}_{a}\sim\mathcal{N}(\mathbf{0},\sigma_{a}^{2}\mathbf{I})$ and $\mathbf{n}_{g}\sim\mathcal{N}(\mathbf{0},\sigma_{g}^{2}\mathbf{I})$. The gravity vector in the global frame is defined as $\mathbf{g}^{G}=[0,\,0,\,g]^{T}$.

Based on the above measurement model, the IMU state evolves according to the continuous-time kinematics
\begin{equation}
\label{eq:continuous_kinematics}
\dot{\mathbf{x}}_{I}(t)=f\!\left(\mathbf{x}_{I}(t),\,\mathbf{a}_{m}^{I}(t),\,\boldsymbol{\omega}_{m}^{I}(t),\,\mathbf{n}(t)\right),
\end{equation}

where $\mathbf{x}_{I}$ denotes the IMU state in \eqref{eq:imu_state}. In implementation, the above nonlinear dynamics are discretized over each IMU sampling interval $\Delta t$ and propagated on the state manifold as
\begin{equation}
\label{eq:discrete_propagation}
\hat{\mathbf{x}}_{i+1}
=
\hat{\mathbf{x}}_{i}
\boxplus
\left(
\Delta t\, f(\hat{\mathbf{x}}_{i},\mathbf{u}_{i},\mathbf{0})
\right),
\end{equation}
where $\mathbf{u}_{i}=\left[(\mathbf{a}_{m,i}^{I})^{T}\;\;(\boldsymbol{\omega}_{m,i}^{I})^{T}\right]^{T}$ denotes the IMU input at the $i$-th sampling instant. The associated covariance is propagated by first-order linearization,
\begin{equation}
\label{eq:covariance_propagation}
\hat{\mathbf{P}}_{i+1}
=
{\mathbf{F}_{x_i}}\hat{\mathbf{P}}_{i}{\mathbf{F}_{x_i}^{T}}
+
\mathbf{F}_{w_i}\mathbf{Q}_{i}\mathbf{F}_{w_i}^{T},
\end{equation}

where $\mathbf{F}_{x_i}$ and $\mathbf{F}_{w_i}$ are the Jacobians of the discrete propagation with respect to the error state and process noise, respectively, and $\mathbf{Q}_{i}$ is the IMU noise covariance. This propagated state serves as the prior estimate for the subsequent LiDAR and visual updates.

To ensure stable inertial propagation from system startup, a short static
segment of approximately $3\,\mathrm{s}$ is used for initialization. During
this interval, the platform is assumed to remain stationary, such that
{the time-averaged accelerometer measurements establish the gravity
direction and vertical axis. The averaged angular-rate direction is used only
to select the horizontal orientation; for a consumer-grade IMU, an
approximately $3\,\mathrm{s}$ average may be dominated by gyroscope bias and
noise and is therefore not interpreted as a reliable absolute-heading
measurement. Together with the corresponding global-frame directions, the
initial triad is completed by the cross-product direction and normalized,
giving}
\begin{equation}
\label{eq:raw_rotation}
\mathbf{C}_{G I_{0}}^{\mathrm{raw}}
=
\left[
\hat{\mathbf{g}}^{I}\;\;
\hat{\boldsymbol{\omega}}^{I}\;\;
\hat{\mathbf{v}}^{I}
\right]
\left[
\hat{\mathbf{g}}^{G}\;\;
\hat{\boldsymbol{\omega}}^{G}\;\;
\hat{\mathbf{v}}^{G}
\right]^{T},
\end{equation}
{where $\hat{\boldsymbol{\omega}}^{I}$ and
$\hat{\boldsymbol{\omega}}^{G}$ select the horizontal axes after gravity
alignment, and $\hat{\mathbf{v}}^{I}$ and $\hat{\mathbf{v}}^{G}$ are their
cross-product directions with gravity.} Since
$\mathbf{C}_{G I_{0}}^{\mathrm{raw}}$ may deviate from $SO(3)$ due to noise
and finite averaging, it is projected onto the closest valid rotation matrix
using singular value decomposition (SVD),
\begin{equation}
\label{eq:svd_projection}
\mathbf{C}_{G I_{0}}^{\mathrm{raw}}
=
\mathbf{U}\mathbf{\Sigma}\mathbf{V}^{T}
\;\;\rightarrow\;\;
\mathbf{C}_{G I_{0}}
=
\mathbf{U}\mathbf{V}^{T},
\end{equation}
and then converted into the quaternion representation
$\hat{\mathbf{q}}_{G I_{0}}$. With the initial attitude fixed, the initial
gyroscope and accelerometer biases are estimated through a small least-squares
problem based on the static IMU measurements, yielding
$\mathbf{b}_{g_{0}}$ and $\mathbf{b}_{a_{0}}$.
{The short interval is not assumed to guarantee a useful absolute
Earth-rate heading. If the angular-rate direction is dominated by bias or
noise, it defines an uncertain but fixed yaw gauge, while gravity still defines
the vertical axis. Because F-LIVO constructs its state and online map in the
same local frame and uses no subsequent absolute-heading observation, a pure
initial yaw offset rotates the local solution without changing relative
odometry. Fixing the initial yaw is therefore equivalent for relative
estimation; a resolved angular-rate direction only provides a more physically
interpretable horizontal orientation. This gauge argument excludes gravity or
IMU-bias errors, which can affect propagation.}
Once the initial attitude and bias states are obtained, the system enters the
normal operating stage, where IMU measurements are continuously used for state
propagation and LiDAR and visual observations are incorporated for subsequent
filter updates.

\begin{figure*}[t]
    \centering
    \subfloat[]{\includegraphics[width=0.31\textwidth]{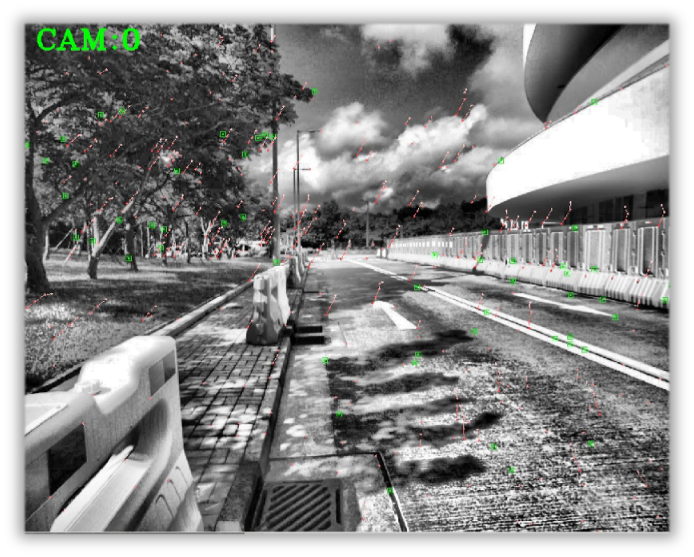}\label{fig:ppt_visual_plane_a}}
    \hfil
    \subfloat[]{\includegraphics[width=0.31\textwidth]{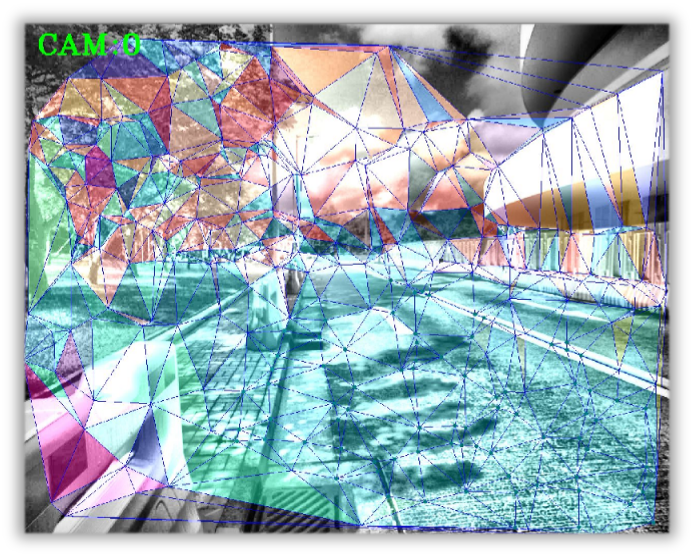}\label{fig:ppt_visual_plane_b}}
    \hfil
    \subfloat[]{\includegraphics[width=0.31\textwidth]{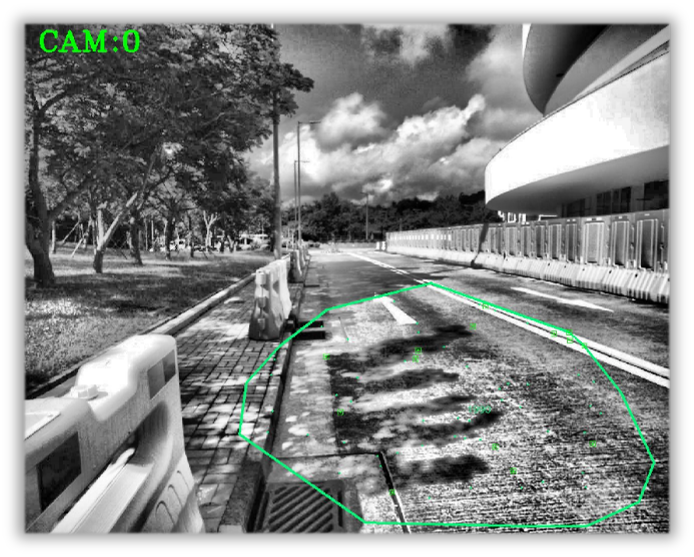}\label{fig:ppt_visual_plane_c}}
    
    \vspace{0.4em}
    \subfloat[]{\includegraphics[width=0.64\textwidth]{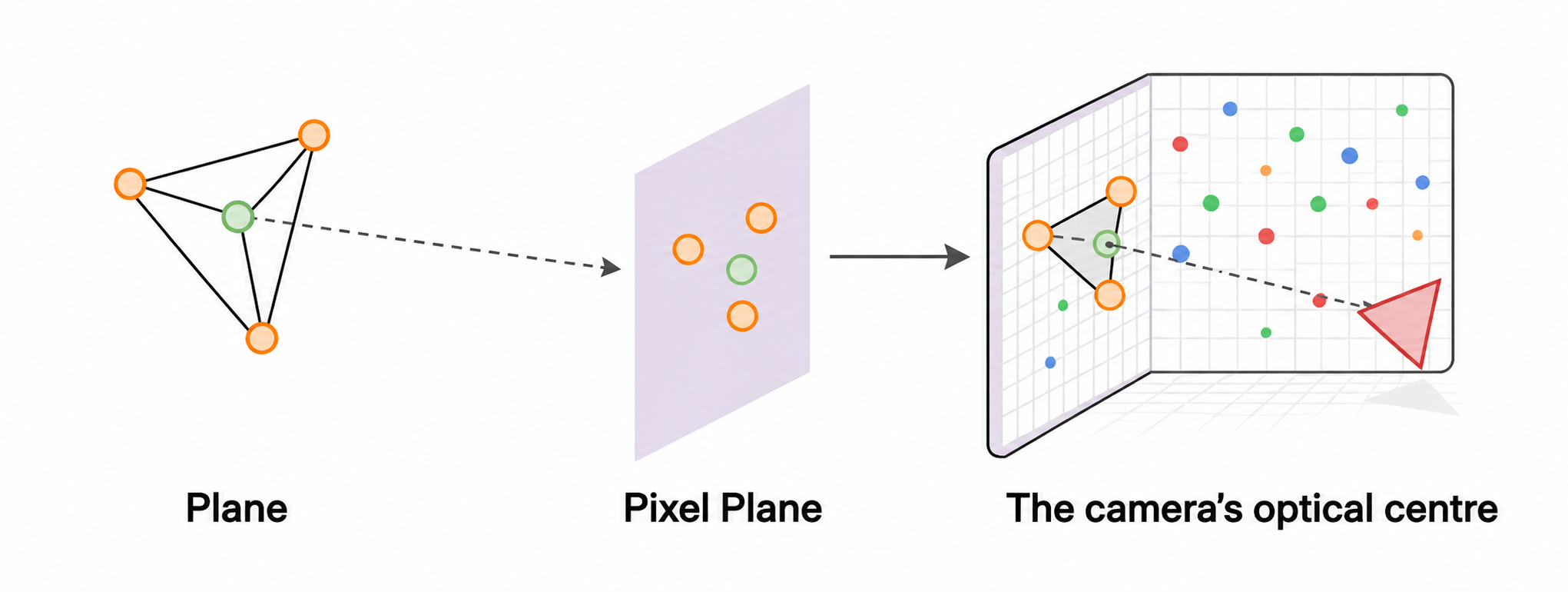}\label{fig:ppt_visual_plane_d}}
    \hfil
    \subfloat[]{\includegraphics[width=0.24\textwidth]{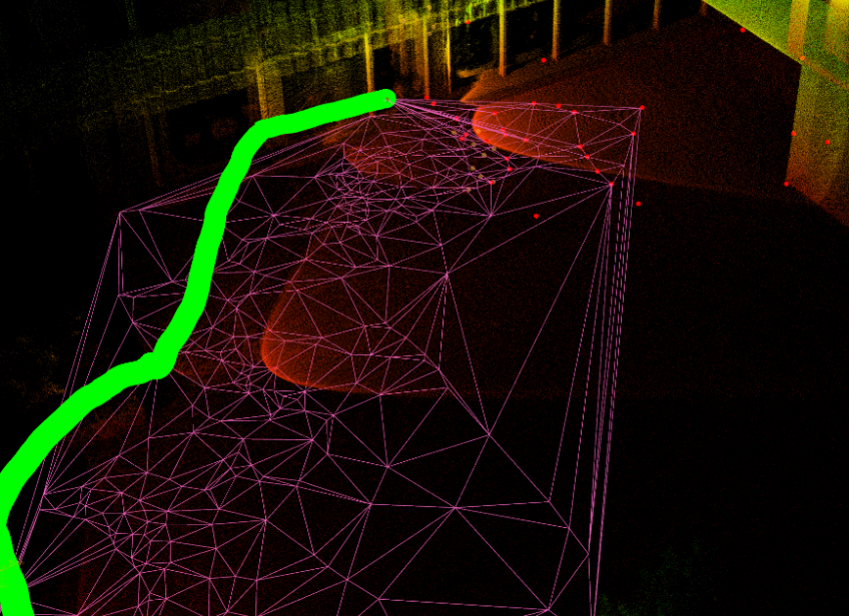}\label{fig:ppt_visual_plane_e}}
    \caption{Illustrations of the visual-plane construction and LiDAR-assisted depth recovery. (a) Triangulated visual feature points. (b) Delaunay triangulation built from the tracked visual features. (c) Coplanar aggregation of neighboring triangles into visual feature planes. (d) LiDAR points projected onto the image plane are used for local search and plane fitting to recover feature depth when visual triangulation is weak. (e) Red points denote visual feature points whose depths are recovered from LiDAR geometry.}
    \label{fig:visual_theory_from_ppt}
\end{figure*}

\subsection{Camera Measurement Update}
The camera measurement update is built upon ov\_plane. On this basis, we further strengthen the link between LIO and VIO at the feature level by introducing cross-modal point-to-plane constraints.

\subsubsection{Structured visual plane construction}
Instead of treating FAST corners as isolated visual landmarks, we organize stable tracked and triangulated features into local triangle patches and then aggregate coplanar patches into visual planes. Let $\mathcal{P}_{f}=\{\mathbf{p}_{f_{i}}^{G}\}_{i=1}^{N_{f}}$ denote the set of valid 3D visual features and $\{\mathbf{u}_{i}\}_{i=1}^{N_{f}}$ their image coordinates. A 2D Delaunay triangulation is performed on the image plane to obtain a stable neighborhood graph,
\begin{equation}
\label{eq:delaunay_graph}
\mathcal{T}=\mathrm{Delaunay}\!\left(\{\mathbf{u}_{i}\}_{i=1}^{N_{f}}\right),
\end{equation}
where each triangle $\tau=(i,j,k)\in\mathcal{T}$ defines a minimal local surface patch. For each triangle, we compute its normal and offset in the global frame as
\begin{equation}
\label{eq:visual_triangle_normal}
\begin{aligned}
\mathbf{v}_{1}^{G}
&=
\frac{\mathbf{p}_{f_{j}}^{G}-\mathbf{p}_{f_{i}}^{G}}
{\left\|\mathbf{p}_{f_{j}}^{G}-\mathbf{p}_{f_{i}}^{G}\right\|},
\quad
\mathbf{v}_{2}^{G}
=
\frac{\mathbf{p}_{f_{k}}^{G}-\mathbf{p}_{f_{i}}^{G}}
{\left\|\mathbf{p}_{f_{k}}^{G}-\mathbf{p}_{f_{i}}^{G}\right\|}, \\
\mathbf{n}_{\tau}^{G}
&=
\frac{\mathbf{v}_{2}^{G}\times\mathbf{v}_{1}^{G}}
{\left\|\mathbf{v}_{2}^{G}\times\mathbf{v}_{1}^{G}\right\|},
\quad
d_{\tau}^{G}
=
(\mathbf{n}_{\tau}^{G})^{T}\mathbf{p}_{f_{i}}^{G}.
\end{aligned}
\end{equation}

Two adjacent triangles $\tau_{a}$ and $\tau_{b}$ are merged into the same visual plane only when they satisfy both normal consistency and point-to-plane consistency:
\begin{equation}
\label{eq:coplanar_merge}
\begin{aligned}
\arccos\!\left(
(\mathbf{n}_{\tau_{a}}^{G})^{T}\,\mathbf{n}_{\tau_{b}}^{G}
\right)
&< \Delta_{\theta}, \\
\left|
(\mathbf{n}_{\tau_{a}}^{G})^{T}\,\mathbf{p}_{f_{m}}^{G}
- d_{\tau_{a}}^{G}
\right|
&< \Delta_{d},
\quad \forall\, m\in\tau_{b}.
\end{aligned}
\end{equation}

Connected triangle patches that pass the above tests are aggregated into a visual plane feature $\boldsymbol{\Pi}^{G}=(\mathbf{n}^{G},d^{G})$. This procedure transforms sparse point tracks into more stable structured planar features, which improves the repeatability of cross-frame and cross-modal associations, as illustrated by the extracted examples in Fig.~\ref{fig:visual_theory_from_ppt}.

\subsubsection{LiDAR-assisted depth recovery}
{LiDAR-assisted depth recovery complements the multi-view visual geometry of a tracked feature rather than acting only as a fallback for failed triangulation. For each observation in a feature track, the LiDAR map can provide a frame-wise depth candidate along the corresponding visual bearing; this support is particularly valuable in low-texture scenes or under weak parallax.} Specifically, each LiDAR map point $\mathbf{p}_{\ell_{j}}^{G}$ is projected onto the current image plane as
\begin{equation}
\label{eq:lidar_projection_camera}
\mathbf{u}_{\ell_{j}}
=
\boldsymbol{\Lambda}\!\left(
\mathbf{R}_{I}^{C}\,
\mathbf{R}_{G}^{I_{k}}\bigl(\mathbf{p}_{\ell_{j}}^{G} - \mathbf{p}_{I_{k}}^{G}\bigr)
+ \mathbf{p}_{I}^{C}
\right),
\end{equation}

and a 2D $k$-d tree is constructed in the pixel domain using the set $\{\mathbf{u}_{\ell_{j}}\}$. For a visual feature observation $\mathbf{u}_{f}$, we search the neighboring projected LiDAR points and fit a local plane $\boldsymbol{\Pi}_{\mathrm{fit}}^{G}=(\mathbf{n}_{\mathrm{fit}}^{G},d_{\mathrm{fit}}^{G})$ using the corresponding 3D LiDAR points. Let $\mathbf{p}_{C_{k}}^{G}$ denote the camera center at time $t_{k}$ and $\mathbf{b}_{f}^{G}$ the unit bearing vector of $\mathbf{u}_{f}$ in the global frame. The recovered 3D point is then obtained from the ray-plane intersection
\begin{equation}
\label{eq:lidar_assisted_depth}
(\mathbf{p}_{f}^{\,\mathrm{lidar}})^{G}
=
\mathbf{p}_{C_{k}}^{G} + \rho_{f}\,\mathbf{b}_{f}^{G},
\qquad
\rho_{f}
=
\frac{d_{\mathrm{fit}}^{G} - (\mathbf{n}_{\mathrm{fit}}^{G})^{T}\,\mathbf{p}_{C_{k}}^{G}}
{(\mathbf{n}_{\mathrm{fit}}^{G})^{T}\,\mathbf{b}_{f}^{G}}.
\end{equation}

{Equation~\eqref{eq:lidar_assisted_depth} gives the LiDAR-supported depth candidate associated with the observation at time $t_k$; it does not define the final landmark independently of the other observations in the track. The fitted plane, camera pose, and image bearing are uncertain, so frame-wise candidates associated with the same tracked feature need not coincide exactly. All observations in the track are instead modeled as measurements of one shared latent landmark $\mathbf{p}_{f}^{G}$, whose position is jointly constrained by the multi-view visual bearings and the accepted LiDAR plane observations in the camera update. If valid planar support or multi-view visual information is persistently absent, this joint construction cannot recover the missing geometric observability.} The projection, neighborhood search, and depth recovery process is illustrated in Fig.~\ref{fig:visual_theory_from_ppt}.

\subsubsection{Cross-modal measurement model}
After the above association stage, the camera update jointly uses four residual blocks: image reprojection of visual points, point-to-visual-plane constraints, point-to-LiDAR-plane constraints, and LiDAR-point-to-visual-plane constraints. The geometric relationship underlying these cross-modal constraints is illustrated in Fig.~\ref{fig:cam_measure}. The camera bearing observations and the point-to-plane distance constraints are used to update the state through the following measurement functions:

\begin{figure}[t]
    \centering
    \includegraphics[width=\linewidth]{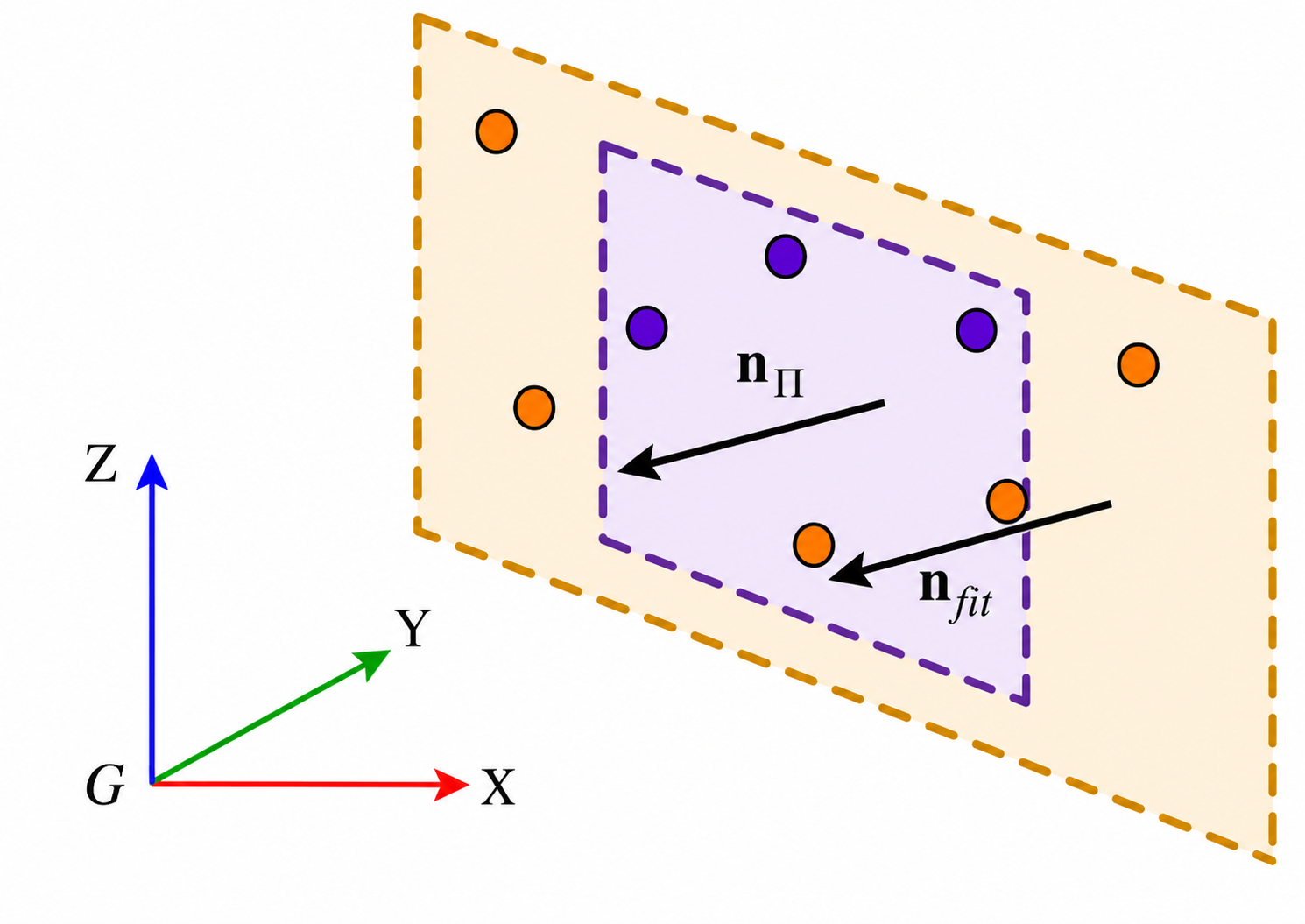}
    \caption{Geometric illustration of the cross-modal coplanar association used in the camera update. Orange points denote LiDAR points and purple points denote visual feature points. When the two sets are associated with the same local plane, point-to-LiDAR-plane and LiDAR-point-to-visual-plane constraints can be constructed consistently.}
    \label{fig:cam_measure}
\end{figure}

\begin{equation}
\label{eq:cam_measurement}
\begin{aligned}
\mathbf{z}_{c}
&= \mathbf{h}_{c}\!\left(\mathbf{x}_{k}\right) + \mathbf{n}_{c} \\
&= \boldsymbol{\Lambda}\!\left(
\mathbf{R}_{I}^{C}\,
\mathbf{R}_{G}^{I_{k}}\bigl(\mathbf{p}_{f}^{G} - \mathbf{p}_{I_{k}}^{G}\bigr)
+ \mathbf{p}_{I}^{C}
\right)
+ \mathbf{n}_{c}
\end{aligned}
\end{equation}

\begin{equation}
\label{eq:plane_measurement}
\mathbf{z}_{d}
= \mathbf{h}_{d}\!\left(\mathbf{x}_{k}\right) + \mathbf{n}_{d}
= \left(
(\mathbf{n}^{G})^{T}\,\mathbf{p}_{f}^{G} - d^{G}
\right)
+ \mathbf{n}_{d}
\end{equation}

where $\boldsymbol{\Lambda}(\cdot)$ denotes the projection from the camera frame to the image plane. $\mathbf{n}_{c}$ and $\mathbf{n}_{d}$ denote zero-mean Gaussian noises for the bearing and point-to-plane measurements, with nominal standard deviations of $1\,\text{pixel}$ and $3\,\text{cm}$, respectively.

{The point $\mathbf{p}_{f}^{G}$ in Eq.~\eqref{eq:plane_fit_measurement} is this shared landmark estimate, not the raw single-frame intersection $(\mathbf{p}_{f}^{\,\mathrm{lidar}})^{G}$ generated by Eq.~\eqref{eq:lidar_assisted_depth}. A raw intersection satisfies its generating plane exactly, but the shared landmark is determined by all visual bearings and accepted plane constraints and is therefore not algebraically forced to satisfy any one fitted plane. Consequently, the residual in Eq.~\eqref{eq:plane_fit_measurement} can be numerically small, but it is not zero by construction.}

To achieve feature-level fusion, we introduce additional point-to-plane constraints between visual features, planar features, and the LiDAR point cloud during the camera measurement update.

\begin{equation}
\label{eq:plane_fit_measurement}
\mathbf{z}_{d}^{\text{fit}}
= \mathbf{h}_{d}\!\left(\mathbf{x}_{k}\right) + \mathbf{n}_{d}
= \left(
(\mathbf{n}_{\text{fit}}^{G})^{T}\,\mathbf{p}_{f}^{G}
- d_{\text{fit}}^{G}
\right)
+ \mathbf{n}_{d}
\end{equation}

\begin{equation}
\label{eq:plane_nearest_measurement}
\mathbf{z}_{d}^{i}
= \mathbf{h}_{d}\!\left(\mathbf{x}_{k}\right) + \mathbf{n}_{d}
= \left(
(\mathbf{n}^{G})^{T}\,(\mathbf{p}_{\text{nearest}}^{\,i})^{G}
- d^{G}
\right)
+ \mathbf{n}_{d}
\end{equation}

These additional residuals explicitly align visual points with LiDAR planes and LiDAR points with visual planes, which yields a unified cross-modal surface representation during the camera update. Although the residuals originate from different sensing modalities, they are all formulated as 3D point-to-plane constraints in the common global frame.

{For each feature track, the visual bearing residuals across the cloned camera poses and the accepted LiDAR plane residuals jointly constrain $\mathbf{p}_{f}^{G}$. Equation~\eqref{eq:lidar_assisted_depth} parameterizes the LiDAR-supported depth relation along an individual visual ray, whereas Eq.~\eqref{eq:plane_fit_measurement} is the corresponding probabilistic plane constraint in the joint update. The depth candidate from Eq.~\eqref{eq:lidar_assisted_depth} is therefore not inserted as an additional independent measurement or prior: each accepted LiDAR plane observation contributes probabilistic information once through the point-to-plane residual. The plane associated with different observations may be a distinct local fit or may represent the same physical surface; statistical independence between such plane fits is not required for the zero-by-construction argument. Projection and ray geometry are evaluated at the current propagated state, while the discrete neighborhood selection and fitted plane are conditioned on the accepted association during one linearized update and are recomputed at the next association step. At first order, the perturbation of the depth candidate can be written as $\delta\mathbf{p}_{f,k}^{\mathrm{lidar}}=\mathbf{J}_{x,k}\delta\mathbf{x}_{k}+\mathbf{J}_{b,k}\delta\mathbf{b}_{f,k}+\mathbf{J}_{\pi,k}\delta\boldsymbol{\pi}_{k}$, where $\delta\boldsymbol{\pi}_{k}$ denotes the fitted-plane perturbation. Because this candidate is not fused as a separate measurement, its induced covariance is not injected a second time; pose and bearing uncertainty enter through the propagated covariance and $\mathbf{n}_{c}$, while the plane-induced component is represented once by the effective point-to-plane noise $\mathbf{n}_{d}$. If several residuals share a fitted plane, $\delta\boldsymbol{\pi}_{k}$ is a common uncertainty component rather than independent information replicated by every feature.}

By stacking the linearized observation equations, we obtain

\begin{equation}
\label{eq:joint_update}
\begin{aligned}
\left[
\begin{array}{c}
\tilde{\mathbf{z}}_{c} \\
\tilde{\mathbf{z}}_{d} \\
\tilde{\mathbf{z}}_{d}^{\mathrm{fit}} \\
\tilde{\mathbf{z}}_{d}^{\,i}
\end{array}
\right]
&=
\left[
\begin{array}{c}
\mathbf{H}_{C}^{c} \\
0 \\
0 \\
0
\end{array}
\right]\tilde{\mathbf{x}}_{C}
+
\left[
\begin{array}{c}
\mathbf{H}_{f}^{c} \\
\mathbf{H}_{f}^{d} \\
\mathbf{H}_{f}^{\mathrm{fit}} \\
0
\end{array}
\right] \tilde{\mathbf{p}}_{f}^{G}
\\[6pt]
&\quad+
\left[
\begin{array}{c}
0 \\
\mathbf{H}_{\pi}^{d} \\
0 \\
\mathbf{H}_{\pi}^{\mathrm{fit}}
\end{array}
\right] \tilde{\Pi}^{G}
+
\left[
\begin{array}{c}
\mathbf{n}_{c} \\
\mathbf{n}_{d} \\
\mathbf{n}_{d} \\
\mathbf{n}_{d}
\end{array}
\right]
\end{aligned}
\end{equation}

where $\tilde{\mathbf{z}}$ is the measurement residual; $\mathbf{H}_{C}$, $\mathbf{H}_{f}$, and $\mathbf{H}_{\pi}$ are the Jacobians for the IMU poses, point feature, and plane feature, respectively.

The features are categorized into two types: MSCKF features and SLAM features. For MSCKF features, the Jacobian is compressed via nullspace projection; see ov\_plane for details. After subsequent noise whitening, the measurement model becomes $\mathbf{n} \sim \mathcal{N}(\mathbf{0},\mathbf{I})$:

\begin{equation}
\label{eq:joint_residual}
\mathbf{z}
= \mathbf{H}_{C}\,\widetilde{\mathbf{x}}_{C}
+ \mathbf{H}_{f}\,\widetilde{\mathbf{p}}_{f}^{G}
+ \mathbf{H}_{\pi}\,\tilde{\boldsymbol{\Pi}}^{G}
+ \mathbf{n}
\end{equation}

Using this linearized measurement model, we can now perform the standard EKF update.

\subsection{LiDAR Measurement Update}
The LiDAR measurement update method references FAST-LIO2, and we further refine both the point-to-plane noise modeling and the IEKF correction process. In our framework, LiDAR-supported planes provide directly observable local geometric statistics, which makes the LiDAR update the natural place to introduce adaptive point-to-plane covariance modeling and efficient iterative state correction.

\subsubsection{Adaptive covariance modeling for point-to-plane residuals}
The measurement model for a point-to-plane constraint in the point cloud is

\begin{equation}
\label{eq:plane_residual_fitted}
\begin{aligned}
\mathbf{z}_{d}
&= \mathbf{h}_{d}\!\left(\mathbf{x}_{k}\right) + \mathbf{n}_{d} \\[4pt]
&=
(\mathbf{n}_{\mathrm{fit}}^{G})^{T}
\biggl(
\mathbf{R}_{I_{k}}^{G}
\Bigl(
\mathbf{R}_{L}^{I}\,\mathbf{p}^{L}
+ \mathbf{p}_{L}^{I}
\Bigr)
+ \mathbf{p}_{I_{k}}^{G}
\biggr)
- d_{\mathrm{fit}}^{G}
+ \mathbf{n}_{d}
\end{aligned}
\end{equation}

In mainstream LIO algorithms, the noise associated with point-to-plane constraints is often modeled by a fixed scalar variance. In practice, however, the reliability of a fitted plane depends on the local surface thickness, the tangential spread of its supporting points, and the number of neighbors used in the fit. To keep the update lightweight while preserving this geometric dependence, we model each point-to-plane residual with an adaptive scalar covariance driven by local plane quality.

For each query point, we retrieve its neighborhood $\mathcal{N}_{\ell}$ and compute the local covariance
\begin{equation}
\label{eq:lidar_local_cov}
\boldsymbol{\Sigma}_{\ell}
=
\frac{1}{K}\sum_{\mathbf{q}_{i}\in\mathcal{N}_{\ell}}
\left(\mathbf{q}_{i}-\bar{\mathbf{q}}\right)
\left(\mathbf{q}_{i}-\bar{\mathbf{q}}\right)^{T}
=
\mathbf{U}\,\mathrm{diag}(\lambda_{3},\lambda_{2},\lambda_{1})\,\mathbf{U}^{T},
\end{equation}

where $\bar{\mathbf{q}}$ is the neighborhood centroid, the eigenvector associated with $\lambda_{1}$ gives the fitted plane normal, and the eigenvalues satisfy $\lambda_{3}\geq\lambda_{2}\geq\lambda_{1}\geq0$. The smallest eigenvalue reflects the surface thickness along the normal direction, while the ratio between the two largest eigenvalues and $\lambda_{1}$ characterizes whether the neighborhood is genuinely planar rather than edge-like or spatially scattered. We therefore define the local planar quality score as

\begin{equation}
\label{eq:local planarity}
\sigma_{\mathrm{plane}}
=
\max\!\left(
\sigma_{\min},
\frac{\lambda_{2}-\lambda_{1}}{\lambda_{3}+\epsilon_{\lambda}}
\right)
\end{equation}

A larger $\sigma_{\mathrm{plane}}$ indicates a more reliable local plane, while a smaller value suggests that the neighborhood is either thick, insufficiently planar, or weakly constrained. Here $\epsilon_{\lambda}$ is a small positive constant used for numerical stability, and $\sigma_{\min}$ is a
lower bound used to avoid unbounded covariance inflation in severely degenerate regions.
{In all experiments, $\sigma_{\min}$ is set to $0.1$, such that $1/\sigma_{\mathrm{plane}}$ is bounded by $10$, and
$\epsilon_{\lambda}$ is set to $10^{-6}\,\mathrm{m}^{2}$ to regularize the denominator when $\lambda_{3}$ approaches zero.}

\begin{figure*}[t]
    \centering
    \subfloat[]{\includegraphics[width=0.29\textwidth]{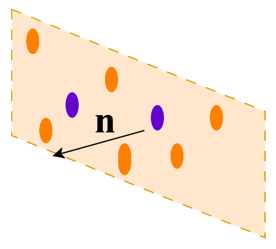}\label{fig:ppt_noise_a}}
    \hfil
    \subfloat[]{\includegraphics[width=0.31\textwidth]{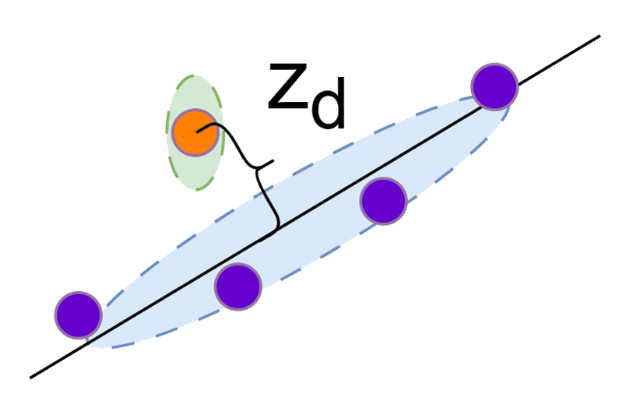}\label{fig:ppt_noise_b}}
    
    \vspace{0.4em}
    \subfloat[]{\includegraphics[width=0.56\textwidth]{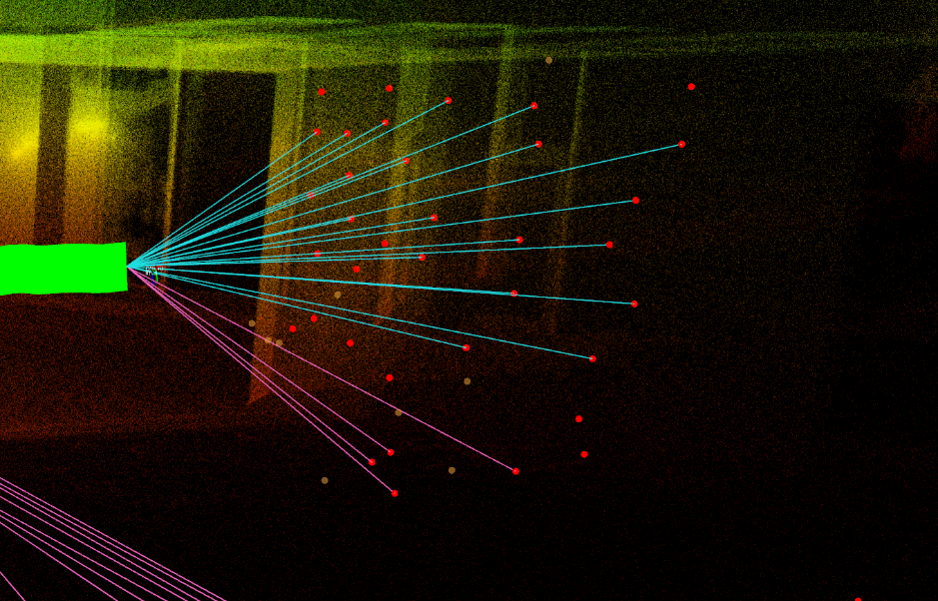}\label{fig:ppt_noise_c}}
    \caption{Illustrations of the adaptive point-to-plane covariance and the unified residual construction. (a) Ideal point-to-plane distance constraint, where the point is assumed to lie on the plane. (b) Practical point-to-plane constraint, where the fitted plane has thickness and the point is generally not exactly on the plane. (c) Unified residual model used for feature-level LiDAR-visual fusion, where the blue and purple rays denote points associated with different plane features.}
    \label{fig:noise_and_residual_from_ppt}
\end{figure*}

Let $\boldsymbol{\Pi}_{\ell}^{G}=(\mathbf{n}_{\ell}^{G},d_{\ell}^{G})$ denote the local fitted plane associated with the current transformed LiDAR point $\mathbf{p}_{\ell}^{G}$, where $d_{\ell}^{G}=(\mathbf{n}_{\ell}^{G})^{T}\bar{\mathbf{q}}_{\ell}^{G}$. The point-to-plane residual can then be written as
\begin{equation}
\label{eq:plane_residual_centroid}
r_{\ell}
=
(\mathbf{n}_{\ell}^{G})^{T}
\left(
\mathbf{p}_{\ell}^{G} - \bar{\mathbf{q}}_{\ell}^{G}
\right),
\end{equation}
which is consistent with equation~\eqref{eq:plane_residual_fitted}. Instead of explicitly propagating every intermediate plane parameter perturbation, we absorb the uncertainty of the transformed point and the fitted plane into a scalar adaptive covariance:

\begin{equation}
\label{eq:nd_noise_model}
\mathbf{R}_{d,\ell}
=
\sigma_{0}^{2}
\left(
1+c_{q}\,
\frac{\lambda_{1}}
{K\,\sigma_{\mathrm{plane}}}
\right),
\end{equation}

where $\sigma_{0}$ denotes the base standard deviation of the point-to-plane residual after motion compensation and $c_{q}>0$ is a fixed scaling coefficient for the dimensionless geometry-dependent covariance multiplier. 
{In all experiments, $\sigma_{\min}=0.1$ and
$\epsilon_{\lambda}=10^{-6}\,\mathrm{m}^{2}$ are used to bound the inverse planar-quality factor and regularize the eigenvalue denominator, respectively. The base LiDAR residual standard deviation is $\sigma_0=0.01$, corresponding to the base covariance $\sigma_0^2=10^{-4}$. The adaptive coefficient is set to $c_q=3.0\times10^{3}\,\mathrm{m}^{-2}$ by requiring the geometry-dependent multiplier to be approximately one for a representative moderately planar neighborhood with $K=5$, $\sqrt{\lambda_1}=0.03\,\mathrm{m}$, and $\sigma_{\mathrm{plane}}=0.5$. For the covariance ablation, the fixed mode applies $\sigma_0=0.01$ uniformly to all valid point-to-plane residuals, whereas the adaptive mode enables the geometry-dependent multiplier in~\eqref{eq:nd_noise_model}. All remaining frontend, association, filter, and map settings are identical. The same covariance parameters are used for all evaluated sequences without sequence-specific tuning.}
This model captures three intuitive effects: a thicker local surface (larger $\lambda_{1}$) increases the residual variance, a better-supported plane fit (larger $K$) reduces the variance, and a weakly planar neighborhood (smaller $\sigma_{\mathrm{plane}}$) is automatically downweighted. Accordingly, each residual is whitened as
\begin{equation}
\tilde{r}_{\ell}
=
\frac{r_{\ell}}{\sqrt{\mathbf{R}_{d,\ell}}},
\qquad
\tilde{n}_{d,\ell}\sim\mathcal{N}(0,1).
\end{equation}
Compared with a fixed covariance model, this formulation preserves the dominant geometric dependence needed for robust filtering while avoiding overly detailed uncertainty propagation that would unnecessarily complicate the update. The geometric interpretation of this adaptive model and its role in the final stacked residuals are shown in Fig.~\ref{fig:noise_and_residual_from_ppt}.

By stacking the linearized observation equations for all valid LiDAR points, the whitened measurement model becomes $\mathbf{n} \sim \mathcal{N}(\mathbf{0},\mathbf{I})$:

\begin{equation}
\label{eq:zd_whitened}
\tilde{\mathbf{z}}_{d}
= \mathbf{H}_{I_{k}}^{d}\,\tilde{\mathbf{x}}_{T_{k}}
+ \mathbf{n}
\end{equation}

\subsubsection{Iterative Extended Kalman Filter (IEKF) update}
{Let $(\hat{\mathbf{x}}_{k\mid k-1},\mathbf{P}_{k\mid k-1})$ denote the prior state and covariance obtained from IMU propagation. Following the standard information-form IEKF update~\cite{gn_iter}, the gain is}

\begin{equation}
\label{eq:kalman_gain_form1}
\mathbf{K}
=
\left(
\mathbf{H}^{T}\mathbf{R}^{-1}\mathbf{H}
+
\mathbf{P}_{k\mid k-1}^{-1}
\right)^{-1}
\mathbf{H}^{T}\mathbf{R}^{-1}
\end{equation}

{Directly applying this expression} to the full state is inefficient because the filter state includes clone poses and mapped features in addition to the current IMU state. In our LiDAR update, however, the point-to-plane residuals directly depend only on the current pose block $\mathbf{x}_{T_k}$. We therefore combine relinearization, QR-based measurement compression, and prior covariance block recovery to perform the IEKF correction on a low-dimensional local error state and then map the result back to the full state.

At the $i$-th IEKF iteration, let $\hat{\mathbf{x}}_{k\mid k}^{(i)}$ be the current iterate and define the local error state and prior displacement on the state manifold as
\begin{equation}
\label{eq:iterative_error_state}
\delta\mathbf{x}_{k}^{(i)}
=
\mathbf{x}_{k} \boxminus \hat{\mathbf{x}}_{k\mid k}^{(i)},
\qquad
\delta\bar{\mathbf{x}}_{k}^{(i)}
=
\hat{\mathbf{x}}_{k\mid k-1} \boxminus \hat{\mathbf{x}}_{k\mid k}^{(i)}.
\end{equation}

Relinearizing the whitened LiDAR residuals around $\hat{\mathbf{x}}_{k\mid k}^{(i)}$ yields
\begin{equation}
\label{eq:iterative_lidar_linearization}
\tilde{\mathbf{r}}_{d}^{(i)}
=
\tilde{\mathbf{H}}_{T_k}^{d,(i)}\,\delta\mathbf{x}_{T_k}^{(i)}
+
\tilde{\mathbf{n}}_{d}^{(i)},
\qquad
\tilde{\mathbf{n}}_{d}^{(i)}\sim\mathcal{N}(\mathbf{0},\mathbf{I}),
\end{equation}

where $\delta\mathbf{x}_{T_k}^{(i)}$ denotes the current-pose component of the error state.

\begin{equation}
\label{eq:measurement_compression}
\begin{aligned}
\left(\mathbf{Q}^{(i)}\right)^{T}\tilde{\mathbf{H}}_{T_k}^{d,(i)}
&=
\begin{bmatrix}
\mathbf{U}^{(i)}\\
\mathbf{0}
\end{bmatrix},
\qquad
\left(\mathbf{Q}^{(i)}\right)^{T}\mathbf{Q}^{(i)}=\mathbf{I}, \\[6pt]
\left(\mathbf{Q}^{(i)}\right)^{T}\tilde{\mathbf{r}}_{d}^{(i)}
&=
\begin{bmatrix}
\breve{\mathbf{r}}_{d}^{(i)}\\
\mathbf{r}_{\perp}^{(i)}
\end{bmatrix}, \\[6pt]
\breve{\mathbf{r}}_{d}^{(i)}
&=
\breve{\mathbf{H}}_{T_k}^{d,(i)}\,\delta\mathbf{x}_{T_k}^{(i)}
+
\breve{\mathbf{n}}_{d}^{(i)},
\qquad
\breve{\mathbf{n}}_{d}^{(i)}\sim\mathcal{N}(\mathbf{0},\mathbf{I}),
\end{aligned}
\end{equation}

where $\mathbf{Q}^{(i)}$ is orthogonal and $\mathbf{U}^{(i)}$ is upper triangular. Because the residuals have already been whitened, the orthogonal transformation preserves the unit covariance of the noise. The lower block $\mathbf{r}_{\perp}^{(i)}$ contains no state-dependent information and can therefore be discarded without changing the posterior; thus we set $\breve{\mathbf{H}}_{T_k}^{d,(i)}=\mathbf{U}^{(i)}$ and use the compressed model for the IEKF correction.

Next, partition the prior covariance and prior displacement according to the current pose block and the remaining states:
\begin{equation}
\label{eq:block_prior_partition}
\mathbf{P}_{k\mid k-1}
=
\begin{bmatrix}
\mathbf{P}_{TT} & \mathbf{P}_{Tr}\\
\mathbf{P}_{rT} & \mathbf{P}_{rr}
\end{bmatrix},
\qquad
\delta\bar{\mathbf{x}}_{k}^{(i)}
=
\begin{bmatrix}
\delta\bar{\mathbf{x}}_{T}^{(i)}\\
\delta\bar{\mathbf{x}}_{r}^{(i)}
\end{bmatrix},
\end{equation}

where the subscript $T$ denotes the current pose block and $r$ collects the remaining states, including the current IMU velocity and biases, clone poses, point features, and plane features.

The compressed LiDAR residual directly constrains only $\delta\mathbf{x}_{T}^{(i)}$. The corresponding local {maximum a posteriori (MAP)} problem is
\begin{equation}
\label{eq:local_pose_map}
\delta\hat{\mathbf{x}}_{T}^{(i)}
=
\arg\min_{\delta\mathbf{x}_{T}}
\left\|
\delta\mathbf{x}_{T}-\delta\bar{\mathbf{x}}_{T}^{(i)}
\right\|_{\mathbf{P}_{TT}^{-1}}^{2}
+
\left\|
\breve{\mathbf{r}}_{d}^{(i)}-\breve{\mathbf{H}}_{T_k}^{d,(i)}\delta\mathbf{x}_{T}
\right\|^{2},
\end{equation}

whose closed-form solution is
\begin{equation}
\label{eq:local_pose_solution}
\begin{aligned}
\mathbf{\Lambda}_{T}^{(i)}
&=
\mathbf{P}_{TT}^{-1}
+
\left(\breve{\mathbf{H}}_{T_k}^{d,(i)}\right)^{T}
\breve{\mathbf{H}}_{T_k}^{d,(i)}, \\[4pt]
\mathbf{g}_{T}^{(i)}
&=
\mathbf{P}_{TT}^{-1}\delta\bar{\mathbf{x}}_{T}^{(i)}
+
\left(\breve{\mathbf{H}}_{T_k}^{d,(i)}\right)^{T}
\breve{\mathbf{r}}_{d}^{(i)}, \\[4pt]
\delta\hat{\mathbf{x}}_{T}^{(i)}
&=
\left(\mathbf{\Lambda}_{T}^{(i)}\right)^{-1}\mathbf{g}_{T}^{(i)}.
\end{aligned}
\end{equation}

Equation~\eqref{eq:local_pose_solution} shows that the only matrix inversion required in each iteration is the small current-pose block $\mathbf{\Lambda}_{T}^{(i)}$, rather than the full system dimension.

The remaining states are updated through the prior cross-covariance:
\begin{equation}
\label{eq:remaining_state_recovery}
\delta\hat{\mathbf{x}}_{r}^{(i)}
=
\delta\bar{\mathbf{x}}_{r}^{(i)}
+
\mathbf{P}_{rT}\mathbf{P}_{TT}^{-1}
\left(
\delta\hat{\mathbf{x}}_{T}^{(i)}
-
\delta\bar{\mathbf{x}}_{T}^{(i)}
\right).
\end{equation}

Equivalently, the corresponding block Kalman gains can be written as
\begin{equation}
\label{eq:block_kalman_gain}
\begin{aligned}
\mathbf{S}^{(i)}
&=
\breve{\mathbf{H}}_{T_k}^{d,(i)}\mathbf{P}_{TT}
\left(\breve{\mathbf{H}}_{T_k}^{d,(i)}\right)^{T}
+
\mathbf{I}, \\[4pt]
\mathbf{K}_{T}^{(i)}
&=
\mathbf{P}_{TT}
\left(\breve{\mathbf{H}}_{T_k}^{d,(i)}\right)^{T}
\left(\mathbf{S}^{(i)}\right)^{-1}, \\[4pt]
\mathbf{K}_{r}^{(i)}
&=
\mathbf{P}_{rT}
\left(\breve{\mathbf{H}}_{T_k}^{d,(i)}\right)^{T}
\left(\mathbf{S}^{(i)}\right)^{-1},
\end{aligned}
\end{equation}

which makes explicit how the local pose correction is mapped back to the full state through the cross-covariance terms.

\begin{equation}
\label{eq:iterative_state_update}
\hat{\mathbf{x}}_{k\mid k}^{(i+1)}
=
\hat{\mathbf{x}}_{k\mid k}^{(i)}
\boxplus
\begin{bmatrix}
\delta\hat{\mathbf{x}}_{T}^{(i)}\\
\delta\hat{\mathbf{x}}_{r}^{(i)}
\end{bmatrix}.
\end{equation}

Thus, each IEKF iteration corresponds to one Gauss--Newton step on the local MAP problem:

\begin{equation}
\label{eq:map_opt}
\min_{\delta\mathbf{x}_{k}^{(i)}}
\left(
\left\|
\delta\mathbf{x}_{k}^{(i)}-\delta\bar{\mathbf{x}}_{k}^{(i)}
\right\|_{\mathbf{P}_{k\mid k-1}^{-1}}^{2}
+
\left\|
\breve{\mathbf{r}}_{d}^{(i)}
-
\breve{\mathbf{H}}^{(i)}\delta\mathbf{x}_{k}^{(i)}
\right\|^{2}
\right)
\end{equation}

with $\breve{\mathbf{H}}^{(i)}=\left[\begin{array}{ll}\breve{\mathbf{H}}_{T_k}^{d,(i)} & \mathbf{0}\end{array}\right]$. The iteration terminates when the increment norm falls below a preset threshold or when the maximum number of iterations is reached. Finally, the covariance is updated at the converged linearization point using the Joseph form
\begin{equation}
\label{eq:joseph_cov_update}
\mathbf{P}_{k\mid k}
=
\left(
\mathbf{I}-\mathbf{K}\breve{\mathbf{H}}
\right)
\mathbf{P}_{k\mid k-1}
\left(
\mathbf{I}-\mathbf{K}\breve{\mathbf{H}}
\right)^{T}
+
\mathbf{K}\mathbf{K}^{T},
\end{equation}

where $\mathbf{K}=\left[(\mathbf{K}_{T})^{T}\;\;(\mathbf{K}_{r})^{T}\right]^{T}$ is evaluated at the final iteration. This update preserves the manifold-consistent state correction while avoiding repeated inversion at the full system dimension.

{

\subsection{Asymmetric Linearization Strategy and Trade-Off}
F-LIVO adopts a unified error-state filtering framework for inertial, visual, and LiDAR measurements. It does not independently estimate a VIO pose and an LIO pose for subsequent pose-level fusion. Let $(\hat{\mathbf{x}}^{-},\mathbf{P}^{-})$ denote the state and covariance after IMU propagation. The camera and LiDAR updates can be abstracted as
\begin{equation}
\label{eq:sequential_visual_lidar_updates}
\begin{aligned}
(\hat{\mathbf{x}}^{v},\mathbf{P}^{v})
&=
\mathcal{U}_{v}
\left(
\hat{\mathbf{x}}^{-},\mathbf{P}^{-};
\mathbf{r}_{v},\mathbf{H}_{v}
\right), \\
(\hat{\mathbf{x}}^{+},\mathbf{P}^{+})
&=
\mathcal{U}_{\ell}
\left(
\hat{\mathbf{x}}^{v},\mathbf{P}^{v};
\mathbf{r}_{\ell},\mathbf{H}_{\ell}
\right),
\end{aligned}
\end{equation}
where $\mathcal{U}_{v}$ is a non-iterative EKF/MSCKF update and $\mathcal{U}_{\ell}$ is an iterative IEKF update. Both updates operate on the same full-state representation and covariance structure, with the posterior state and covariance of the camera update serving as the prior for the subsequent LiDAR update. Therefore, the system does not fuse independently estimated VIO and LIO poses at the pose level.

{The standard equivalence between sequential and stacked Kalman updates holds only for a fixed linear-Gaussian model with block-diagonal measurement covariance. This identity is included solely as a linear reference and is not asserted for the nonlinear F-LIVO implementation. Both measurement blocks are processed within the same error-state estimator, but their linearization schedules differ. Because they may be constructed from overlapping local geometry and the map is formed using previous state estimates, strict statistical independence of the effective residual errors is not claimed. The retained-state covariance is propagated between processed measurement blocks, whereas possible cross-block correlation induced through the conditioned geometry is not explicitly represented. Consequently, the idealized equivalence is not used as a consistency guarantee for F-LIVO, and the omitted correlation may affect covariance consistency.}

The visual reprojection and cross-modal point-to-plane measurements are
linearized once around the current predicted state. Defining the stacked visual
innovation as
$\tilde{\mathbf{z}}_{v}
=
\mathbf{z}_{v}
-
\mathbf{h}_{v}(\hat{\mathbf{x}}^{-})$,
the corresponding first-order measurement model is

\begin{equation}
\label{eq:visual_single_linearization}
\tilde{\mathbf{z}}_{v}
\approx
\mathbf{H}_{v}\,\delta\mathbf{x}
+
\mathbf{n}_{v}.
\end{equation}

For MSCKF features, temporary feature-position errors are removed by nullspace projection, whereas SLAM features and plane features maintained in the filter are represented by the corresponding blocks of $\mathbf{H}_{v}$. This one-step formulation avoids repeated visual projection, geometric association, nullspace projection, and measurement compression, thereby keeping the sparse camera update efficient. Its accuracy, however, depends on the quality of the visual linearization point.

{Accordingly, the accepted visual/cross-modal associations, Jacobians, and nominal measurement covariances are held fixed within one camera update.}

The LiDAR update is iterated because its scan-to-map measurements depend
more strongly on the current state. At the $i$-th iteration, defining the
LiDAR innovation as
$\tilde{\mathbf{z}}_{\ell}^{(i)}
=
\mathbf{z}_{\ell}
-
\mathbf{h}_{\ell}^{(i)}
(\hat{\mathbf{x}}^{(i)})$,
the relinearized measurement model is

\begin{equation}
\label{eq:lidar_iterative_relinearization}
\tilde{\mathbf{z}}_{\ell}^{(i)}
\approx
\mathbf{H}_{\ell}^{(i)}
\delta\mathbf{x}^{(i)}
+
\mathbf{n}_{\ell}^{(i)}.
\end{equation}

{The transformed points, nearest-neighbor associations, fitted local planes, residuals, Jacobians, and geometry-dependent whitening factors determined by the effective point-to-plane noise covariances are recomputed after the state changes.} Iterative relinearization is therefore concentrated on the LiDAR constraints, for which pose-dependent association is most significant.

{This asymmetric linearization schedule} represents an accuracy--efficiency trade-off. It does not eliminate the linearization error of the non-iterative visual update. Rather, this error is generally limited when the IMU/LiDAR prior is sufficiently accurate and the visual feature and plane associations remain valid. The approximation may become less accurate under large state errors, aggressive motion, temporal misalignment, low parallax, incorrect feature tracking, or unreliable plane association. These effects are mitigated by IMU propagation, frequent LiDAR IEKF corrections, first-estimate Jacobians (FEJ), chi-square gating, fallback checks, and feature rejection.

{A controlled comparison among the current one-step visual update, an iterated visual update with reassociation or relinearization, and a consistently iterated joint visual--LiDAR update is a valuable extension. Such a comparison requires additional estimator variants with joint association, relinearization, and correlated-noise handling and is left for future work.}

}

\subsection{Map Management}
In this work, the map is organized using an incremental {$k$-dimensional tree (iKD-Tree)}\cite{ikdtree}, which enables efficient incremental updates and stable nearest-neighbor queries under real-time constraints. By maintaining an approximately balanced structure during point insertion and replacement, the iKD-Tree avoids the performance degradation or periodic reconstruction commonly encountered in static KD-Tree methods, thereby ensuring consistent query efficiency during continuous operation.

{Some measurement-level LIVO implementations maintain additional per-point attributes to support cross-modal association. For example, FAST-LIVO2 associates selected mapped LiDAR points with image patches for photometric residual construction, while R3LIVE stores appearance-related information for photometric consistency. These implementation choices incur persistent cross-modal map storage, association, and maintenance costs, but they do not imply that all measurement-level methods use the same representation or are inherently inefficient.}

{F-LIVO organizes its shared geometric map using an iKD-Tree and does not attach persistent dense photometric patches or appearance attributes to mapped LiDAR points. This simplifies persistent cross-modal map maintenance; the complete process-level resource difference from other implementations is evaluated separately and is not attributed solely to this map component.}

{F-LIVO does not discard visual features, but avoids maintaining persistent dense photometric or appearance attributes on mapped LiDAR points. Sparse MSCKF features, SLAM features, structured visual planes, visual reprojection residuals, and cross-modal geometric constraints are retained for state estimation. This design simplifies cross-modal map storage and maintenance, but may provide fewer dense photometric and long-baseline visual constraints in long-horizon scenes with rich texture and weak or repetitive LiDAR geometry, representing a potential accuracy--complexity trade-off.}

\section{EXPERIMENT}
In this section, we evaluate the proposed method on the HILTI SLAM Challenge 2021 dataset\cite{hilti} and the i2Nav-Robot dataset\cite{i2navrobot}, where accuracy is assessed using evo\cite{evo}. {We then use the HKU Mars-Lab dataset to assess estimator continuity under the specific sensor and geometric-observability degradations considered in this work.} Finally, we report computational efficiency and system occupancy. The proposed method is compared with several representative state-of-the-art LIVO systems, including FAST-LIVO2\cite{fastlivo2}.

\begin{figure}[h]
    \centering
    \includegraphics[width=\linewidth]{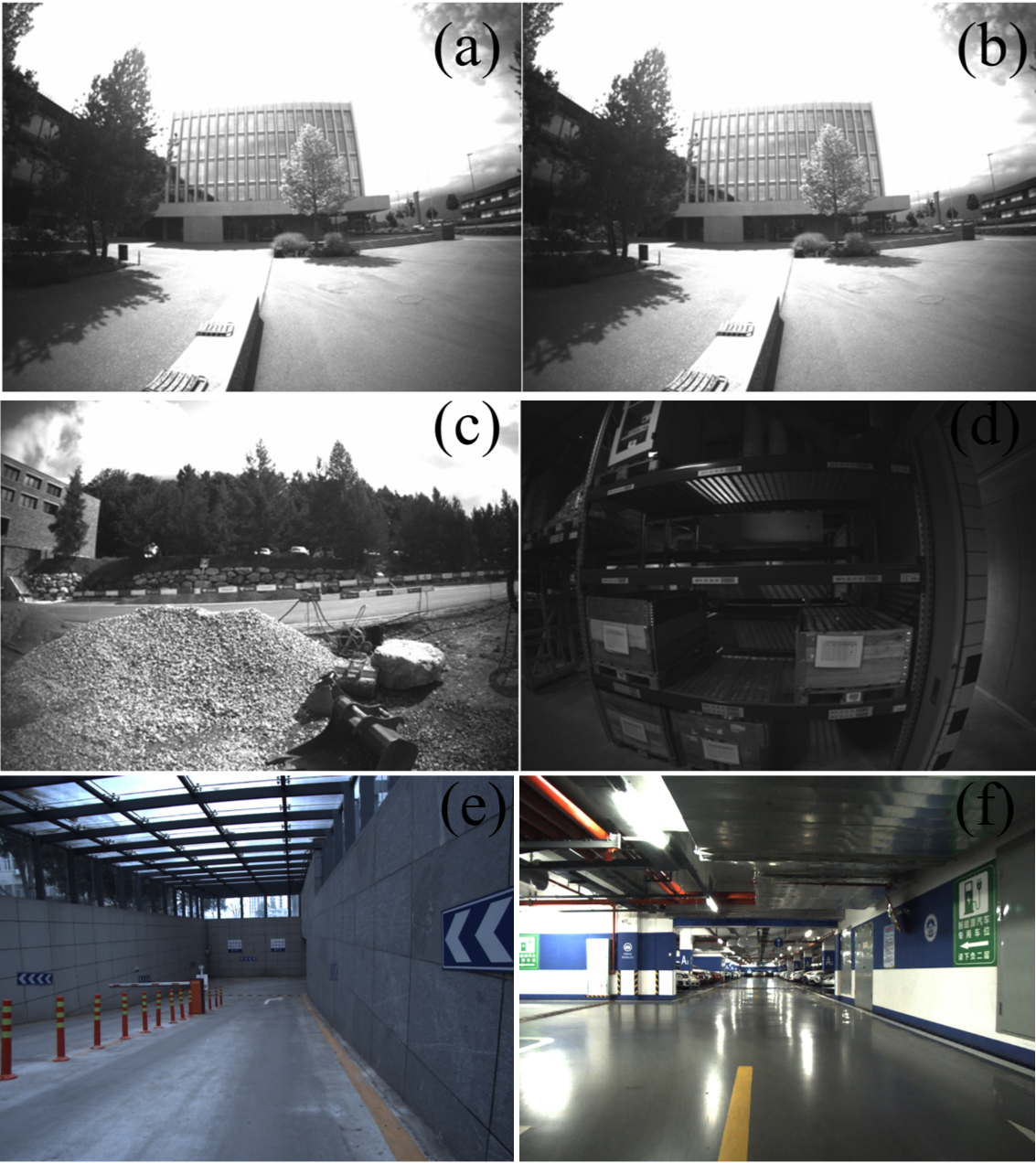}
    \caption{(a) is the construction environment, rich in planar structure. 
    (b) shows the campus environment, filled with piles of concrete and gravel. 
    Both (a) and (b) have overexposure. 
    (c) and (d) are basement environments with narrow passages and poor lighting conditions.
    (e) and (f) correspond to long-sequence scenarios. 
    }
    \label{fig:dataset_env}
\end{figure}

All experiments were conducted on a {desktop workstation} with an AMD Ryzen~5~3600X CPU and 32\,GB {random-access memory (RAM)}.

{The CPU and RSS statistics provide a process-level comparison of the evaluated implementations under the same hardware platform, sensor input, analysis interval, and profiling procedure. They characterize the complete tested processes under their selected configurations, rather than component-isolated map or storage effects, embedded deployment, or hardware-independent performance.}

{The absolute resource values may vary with processor architecture and LiDAR characteristics such as point density, scanning pattern, field of view, and measurement rate; therefore, the reported measurements apply to the evaluated hardware and sensor configuration.}

{For all comparative experiments, each method processes the same sensor data over the same sequence interval using the provided calibration and a common synchronization policy. The parameters of each implementation are kept fixed across sequences, without sequence-specific retuning.} {For resource profiling, the common hardware, sensor input, analysis interval, and monitoring procedure are held fixed; the reported values nevertheless include the method-specific internal map, buffering, and output-related data structures retained by each complete process under its selected configuration.}

\subsection{Accuracy experiment}
To rigorously evaluate positioning accuracy, eight representative sequences were selected from the HILTI SLAM Challenge 2021 dataset (\url{https://hilti-challenge.com/dataset-2021.html}) and i2Nav-Robot dataset (\url{https://github.com/i2Nav-WHU/i2Nav-Robot}), each providing precise {three-degree-of-freedom (3-DoF)} ground-truth references. The sensor suite comprises an Alphasense forward-looking monocular camera, a Livox MID-70 LiDAR, and a consumer-grade IMU.

{The public implementations of R2LIVE and R3LIVE could not initialize correctly under the inverted-IMU configuration and non-identity sensor extrinsics of the evaluated datasets. We therefore made only the minimum adaptations required for initialization and extrinsic compatibility. Their state definitions, measurement residuals, fusion architectures, optimization procedures, and map representations remain unchanged.}

We compare F-LIVO with several representative algorithms known for strong practical performance: R3LIVE, R2LIVE, and FAST-LIVO2. Because FAST-LIVO2 and our method exhibit very similar accuracy, we include a dedicated comparison with FAST-LIVO2 to ensure a rigorous and objective evaluation.

{Among these methods, R2LIVE combines modality-specific visual reprojection and LiDAR geometric residuals within a shared estimator, whereas R3LIVE and FAST-LIVO2 establish direct visual--LiDAR coupling primarily through photometric residuals. FAST-LIO2 is a tightly-coupled LiDAR--inertial system without visual integration, and ov\_plane implements point-and-plane VIO without LiDAR. For completeness, we also conduct ablation experiments using the original FAST-LIO2 and ov\_plane subsystems. {Here, ``w/'' and ``w/o'' denote ``with'' and ``without'', respectively.} ``F-LIVO (w/)'' denotes our full feature-level fusion approach, whereas ``F-LIVO (w/o)'' removes the proposed cross-modal point-to-plane residuals while retaining the modality-specific visual and LiDAR updates in the shared estimator.}

{The root-mean-square error (RMSE) and maximum error (MAX) of the absolute trajectory error (ATE) are reported below.}

\begin{table*}[t]
\centering
\renewcommand{\arraystretch}{1.25} 
\caption{The RMSE/MAX of ATE (UNIT: cm) results in HILTI and i2Nav-Robot dataset. 
The best results are marked in bold.}
\label{tab:hilt_rmse}

\begin{tabular}{lccccccc}
\hline
 & \textbf{R3LIVE} 
 & \textbf{R2LIVE} 
 & \textbf{FAST-LIVO2} 
 & \textbf{FAST-LIO2} 
 & \textbf{ov\_plane} 
 & \textbf{F-LIVO (w/)} 
 & \textbf{F-LIVO (w/o)} \\
\hline

\textit{Basement1}      
 & 17.5/31.2 
 & 19.9/29.5 
 & \textbf{15.8}/26.6 
 & 15.8/27.0 
 & --/-- 
 & \textbf{15.8}/28.0 
 & 15.9/28.3 \\

\textit{Basement4}      
 & 73.6/181.3 
 & 10.0/20.9 
 & {9.40/17.8}
 & 10.6/25.3 
 & 50.5/136.0 
 & {\textbf{6.04}/11.5}
 & 57.2/126.1 \\

\textit{Campus2}        
 & 6.2/10.7 
 & 5.2/10.2 
 & \textbf{4.2}/10.1 
 & 7.0/16.8 
 & 62.4/90.1 
 & 5.7/9.7 
 & 8.0/17.7 \\

\textit{Construction2}  
 & 9.7/\textbf{19.8} 
 & 10.0/20.9 
 & 9.8/\textbf{13.9} 
 & 12.4/26.5 
 & 261.3/1054.1 
 & \textbf{8.2}/25.1 
 & 20.2/28.4 \\

\textit{LabSurvey2}     
 & --/-- 
 & --/-- 
 & --/-- 
 & --/-- 
 & --/-- 
 & \textbf{5.5}/25.8 
 & --/-- \\

\textit{UzhArea2}       
 & \textbf{22.8}/38.7 
 & 26.4/43.5 
 & 23.6/39.7 
 & 29.8/48.5 
 & --/-- 
 & 23.5/42.4 
 & 26.8/46.0 \\

\textit{Parking}
 & \textbf{16.0}/98.3
 & --/--
 & 28.2/83.5
 & 66.7/244.1
 & --/--
 & 55.2/218.2
 & 98.6/249.5 \\

\textit{Building}
 & --/--
 & --/--
 & --/-- 
 & 36.3/135.7
 & --/--
 & \textbf{34.6}/99.2    
 & --/-- 
 \\

\hline
\end{tabular}
\end{table*}

\begin{figure}[t]
    \centering
    \includegraphics[width=\linewidth]{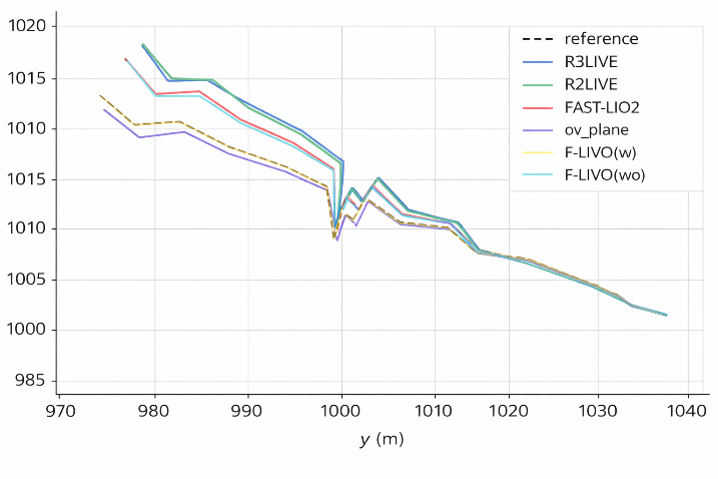}
    \caption{{Representative trajectory estimates on \textit{Basement4} for the compared methods. The repeated-evaluation ATE RMSE/MAX is reported in Table~\ref{tab:hilt_rmse}.}}
    \label{fig:evo_traj_all}
\end{figure}

\begin{figure}[t]
    \centering
    \includegraphics[width=\linewidth]{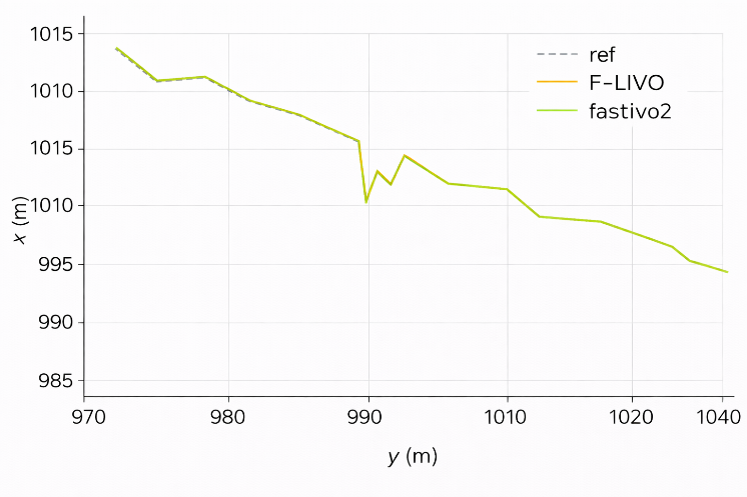}
    \caption{{Representative trajectory estimates for F-LIVO and FAST-LIVO2 on \textit{Basement4}. The repeated-evaluation ATE RMSE/MAX is reported in Table~\ref{tab:hilt_rmse}.}}
    \label{fig:evo_detail}
\end{figure}

As shown in Table~\ref{tab:hilt_rmse}, F-LIVO (w/) achieves competitive accuracy across most evaluated sequences. {On \textit{Basement4}, F-LIVO achieves an ATE RMSE of 6.04~cm and a MAX ATE of 11.5~cm, compared with 9.40~cm and 17.8~cm for FAST-LIVO2 under the repeated evaluation. This result indicates that F-LIVO maintains lower translational error on this sequence.} \textit{Construction2} and \textit{Basement4} both limit the informativeness of visual and LiDAR measurements. \textit{Construction2} contains repetitive structural patterns and cluttered construction materials that reduce visual feature distinctiveness, whereas \textit{Basement4} is a narrow, shelf-dominated warehouse environment that produces highly anisotropic LiDAR geometry and frequent occlusions. {The trajectory plots in Figs.~\ref{fig:evo_traj_all} and~\ref{fig:evo_detail} provide qualitative views of representative estimates on \textit{Basement4}; the quantitative comparison is based on the repeated-evaluation ATE RMSE/MAX reported in Table~\ref{tab:hilt_rmse}.}

{The comparison between F-LIVO (w/) and F-LIVO (w/o) in Table~\ref{tab:hilt_rmse} evaluates the contribution of the proposed feature-level visual--LiDAR observation constraints. Both configurations retain the same visual, LiDAR, and inertial estimation framework, while F-LIVO (w/o) removes the cross-modal feature observation constraints. The limited difference on \textit{Basement1} indicates that their contribution is small when the original measurements are already sufficiently informative. In contrast, the larger improvements on \textit{Basement4}, \textit{Construction2}, \textit{Parking}, \textit{LabSurvey2}, and \textit{Building} show that these constraints become more important under weak or incomplete modality-specific observations.}

\begin{figure}[h]
    \centering
    \includegraphics[width=\linewidth]{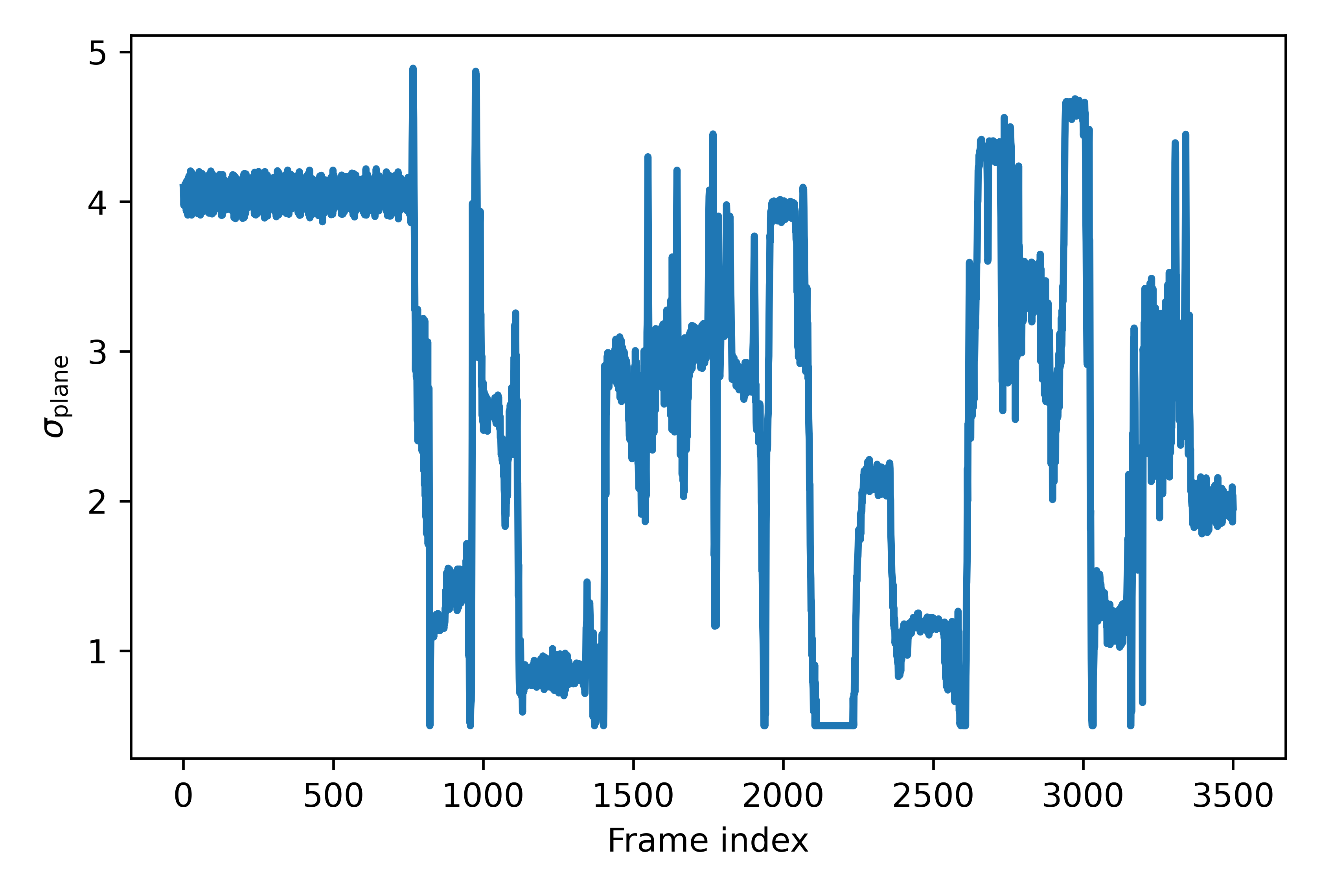}
    \caption{Evolution of the planar quality factor $\sigma_{\mathrm{plane}}$ on the \textit{Basement4} sequence.}
    \label{fig:sigma_plane}
\end{figure}

The behavior observed in these sequences is closely tied to the underlying measurement characteristics. The adaptive covariance formulation in our LIO subsystem scales each point-to-plane residual according to local plane thickness, neighborhood support, and planarity, {thereby reducing the risk of overconfident updates caused by weak or poorly supported local geometry}. This mechanism is crucial in the stationary interval of \textit{Basement4}, where the platform remains motionless in front of shelving units and LiDAR returns exhibit predominantly planar structure. Fig.~\ref{fig:sigma_plane} shows the evolution of the planar quality factor $\sigma_{\mathrm{plane}}$ on \textit{Basement4}. This variation illustrates how the covariance model adjusts the
weight of local planar observations during the stationary interval of Basement4. By contrast, in well-conditioned sequences such as \textit{Campus2} and \textit{UzhArea2}, which contain ample geometric and visual diversity, F-LIVO (w/) closely matches the accuracy of FAST-LIVO2, demonstrating that the proposed fusion strategy introduces negligible overhead or bias when the environment provides sufficiently rich measurements.

{It should be noted that the proposed adaptive covariance evaluates the local geometric fitting quality of an accepted point-to-plane association rather than the physical validity of the LiDAR range measurement. Glass, specular reflection, or multipath propagation may produce a geometrically thin and well-supported but physically incorrect surface. Such a surface may still have a small normal-direction eigenvalue and a high planarity score, resulting in an underestimated covariance. The current association checks, residual gating, and iterative reassociation can mitigate measurements that are inconsistent with the estimated geometry, but they cannot guarantee the rejection of a coherent yet physically incorrect plane. Detecting these failures requires additional sensor-level or temporal cues, such as return intensity, multiple-return information, temporal depth consistency, or explicit reflection modeling.}

In \textit{Construction2}, our accuracy is slightly lower than that of R3LIVE, consistent with their algorithmic differences. The repetitive scaffolding and texture ambiguity in this sequence impair visual tracking and triangulation effectiveness; under such conditions, the denser visual--LiDAR associations in R3LIVE may retain an
advantage under this particular condition. Nevertheless, F-LIVO (w/) maintains centimeter-level accuracy and substantially outperforms pure LIO and pure VIO baselines, illustrating a controlled and graceful degradation as visual constraints diminish rather than a sharp collapse in performance.

{On the long-horizon \textit{Parking} and \textit{Building} sequences, each close to 1.5~km and over 20~minutes in duration, F-LIVO (w/) obtains ATE RMSE/MAX values of 55.2/218.2~cm and 34.6/99.2~cm, respectively. On \textit{Parking}, FAST-LIVO2 achieves a substantially lower RMSE/MAX of 28.2/83.5~cm, while R3LIVE obtains 16.0/98.3~cm. The RMSE of F-LIVO is therefore approximately 1.96 times that of FAST-LIVO2, showing that the proposed formulation does not provide a universal accuracy advantage on long-duration trajectories.}

{The accuracy gap on \textit{Parking} is consistent with the extended intervals of limited motion excitation and low visual parallax observed in this sequence. These conditions reduce the contribution of the sparse visual geometric constraints and cause F-LIVO to rely more heavily on LiDAR--inertial information. The denser visual--LiDAR associations employed by FAST-LIVO2 may retain additional useful constraints under these conditions, which is consistent with its lower trajectory-wide ATE RMSE. Because the present comparison does not isolate this design choice from the other algorithmic differences between the methods, we do not attribute the observed gap to a single component. This result represents an important accuracy limitation of the current feature-level formulation.}

\begin{figure*}[!t]
    \centering
    \subfloat[{Process CPU usage.}]{
        \includegraphics[width=0.47\textwidth]
        {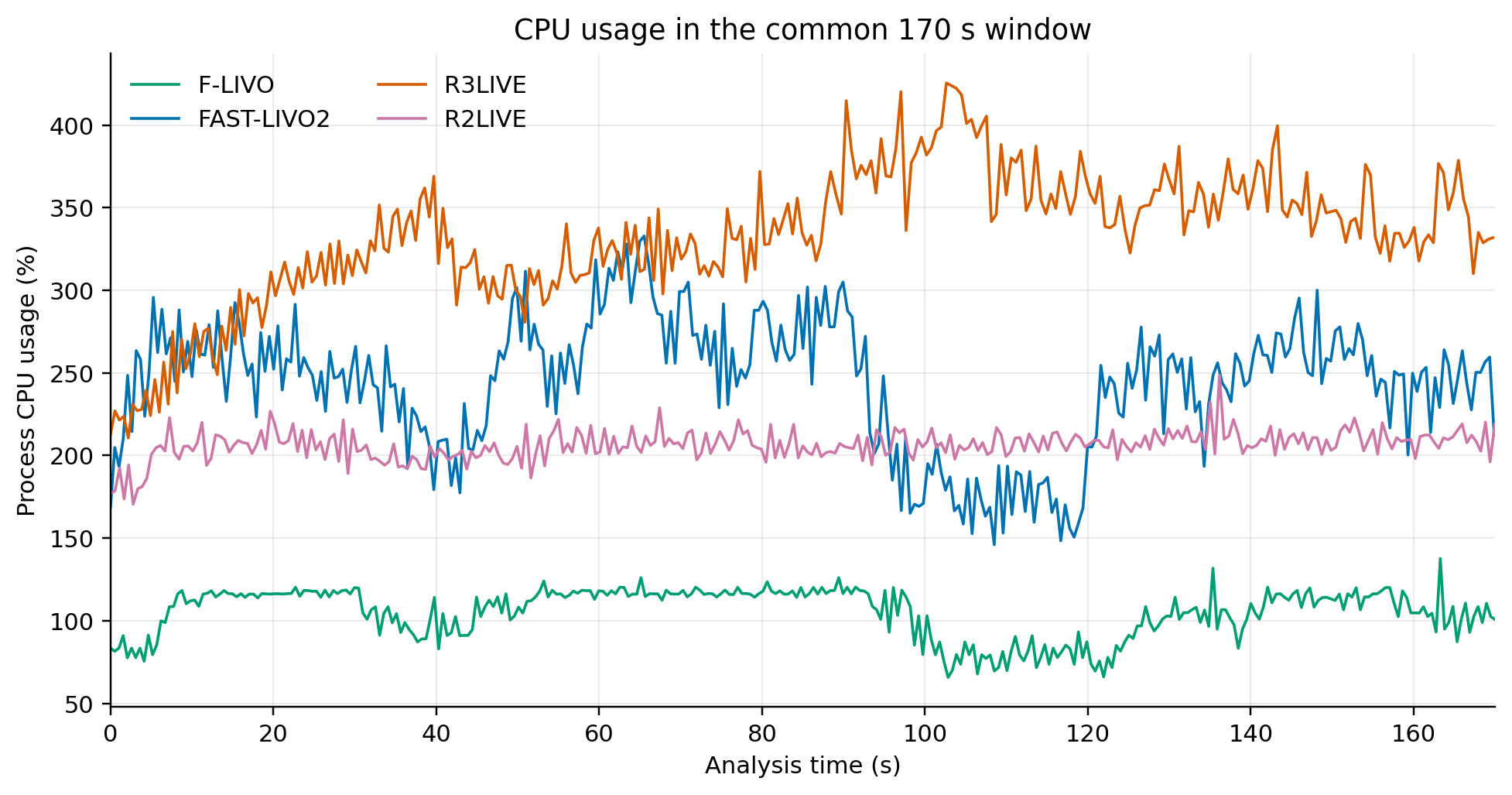}
    }%
    \hfill
    \subfloat[{Process memory usage.}]{
        \includegraphics[width=0.47\textwidth]
        {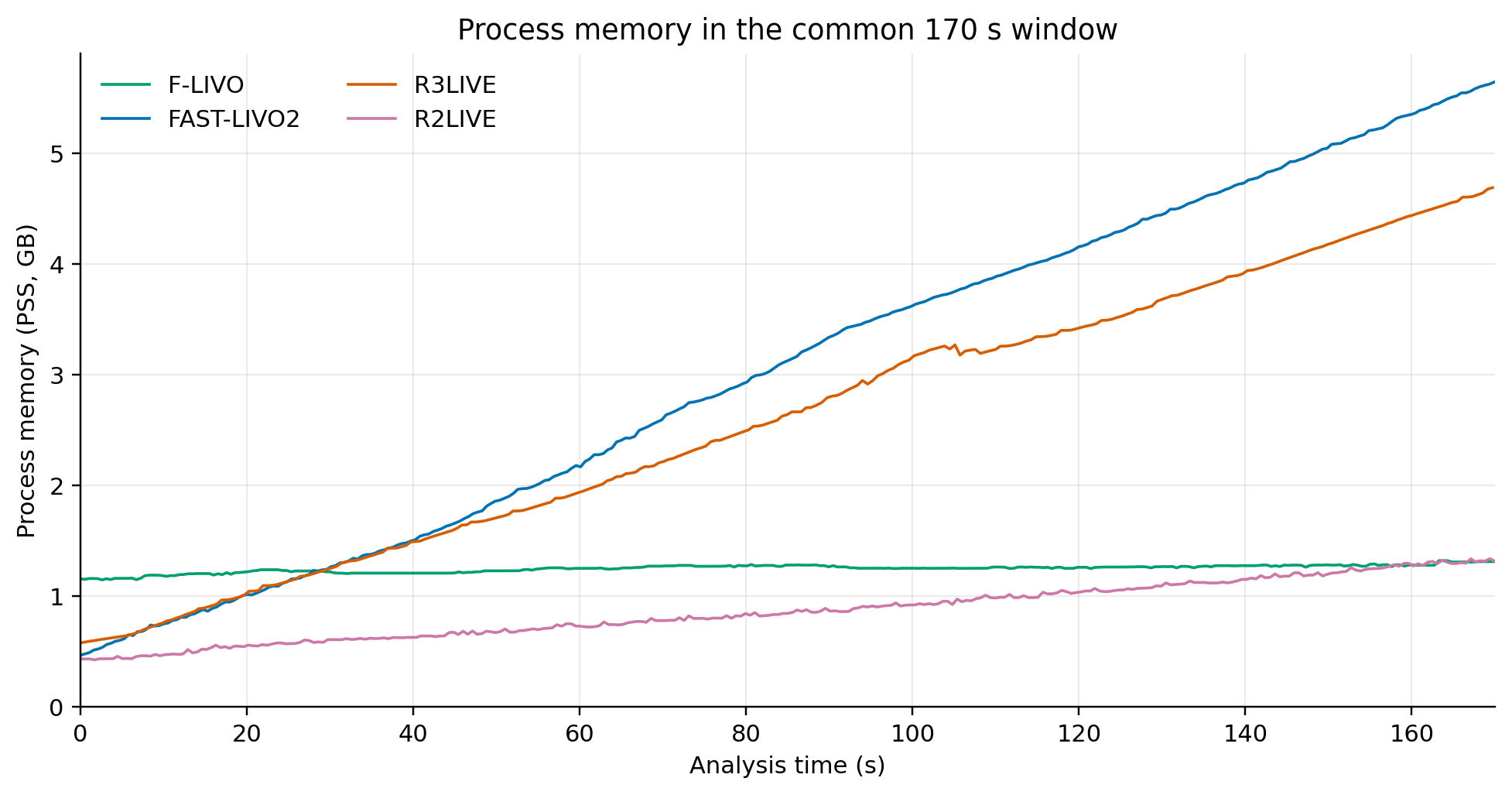}
    }
    \caption{{Process-level resource usage on
    \texttt{hku\_campus\_seq\_00} under the unified profiling protocol.
    (a) Aggregate process CPU usage, where 100\% corresponds to one logical
    core. (b) Resident set size (RSS). All methods are evaluated over the same
    170-s interval after excluding the first 10~s for initialization.}}
    \label{fig:system_occupancy}
\end{figure*}

{The resource results in Fig.~\ref{fig:system_occupancy} and Table~\ref{tab:system_occupancy} provide a separate process-level comparison of the complete tested processes on \texttt{hku\_campus\_seq\_00} under their selected configurations. Because the accuracy and resource measurements are obtained on different sequences, they are treated as complementary evaluations rather than as a controlled same-sequence accuracy--resource Pareto analysis. Table~\ref{tab:system_occupancy} does not isolate the resource effect of an individual map representation, fusion strategy, or output component.}

{All reported comparisons use the fixed per-method configurations described above. Methods that do not produce a valid complete trajectory under these configurations are marked with a dash; results from separately reduced or sequence-specific configurations are not included. Under this protocol, F-LIVO (w/) maintains a valid complete trajectory on \textit{Building} with an ATE RMSE of 34.6~cm, whereas several comparison implementations do not produce valid complete results. This result supports long-horizon estimator continuity, but it does not remove the accuracy limitation observed on \textit{Parking}.}

\begin{figure}[h]
    \centering
    \includegraphics[width=\linewidth]{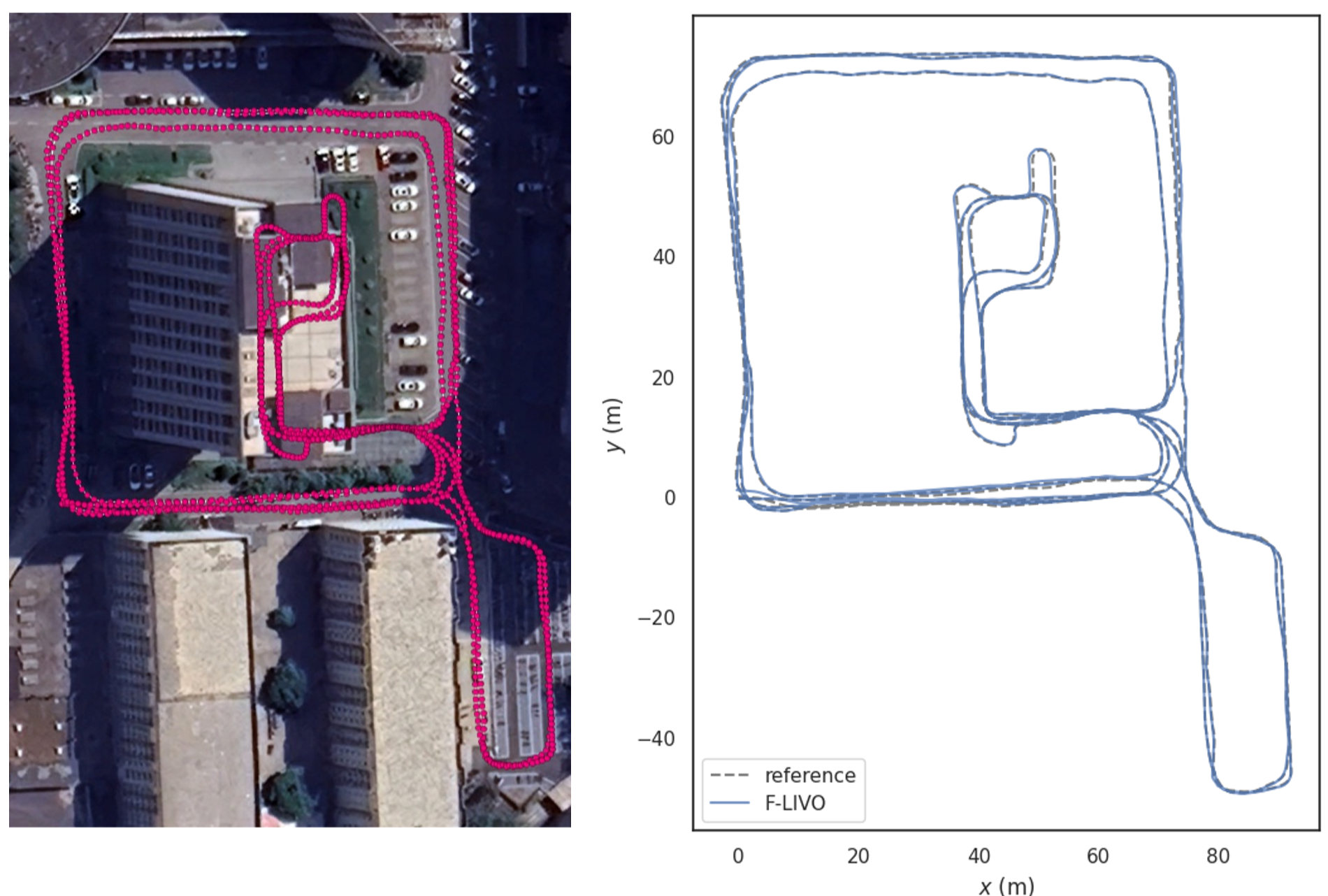}
    \caption{F-LIVO's trajectory in \textit{Building}.}
    \label{fig:traj}
\end{figure}

\subsection{Degeneration experiment}

\begin{figure}[t]
    \centering
    \includegraphics[width=\linewidth]{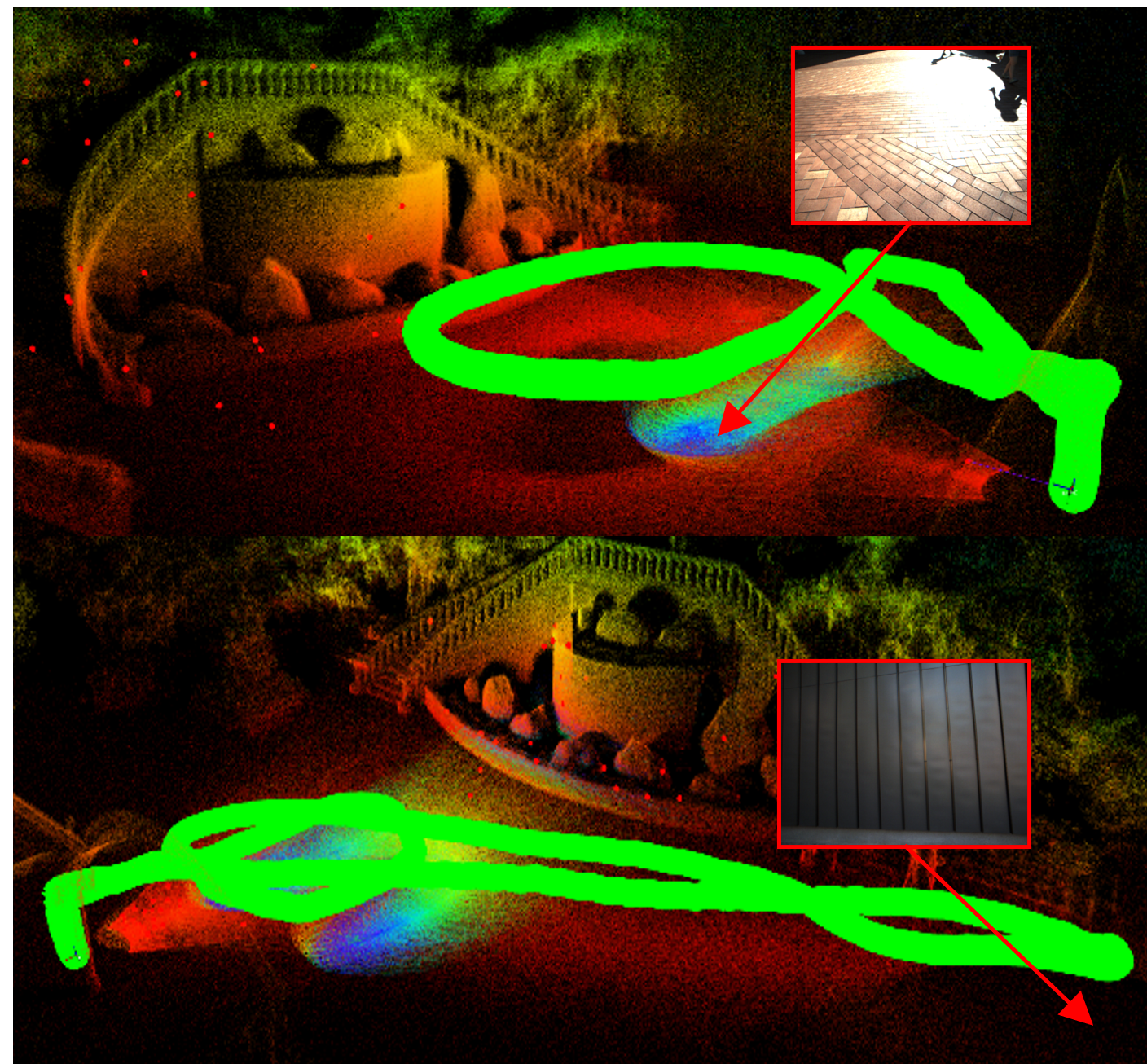}
    \caption{{Results on degenerate sequences~01 and~02 show that F-LIVO maintains continuous estimation with comparatively low accumulated end-to-end drift under the evaluated concurrent visual and LiDAR degradation.}}
    \label{fig:de_seq}
\end{figure}

\begin{figure}[t]
    \centering
    \includegraphics[width=\linewidth]{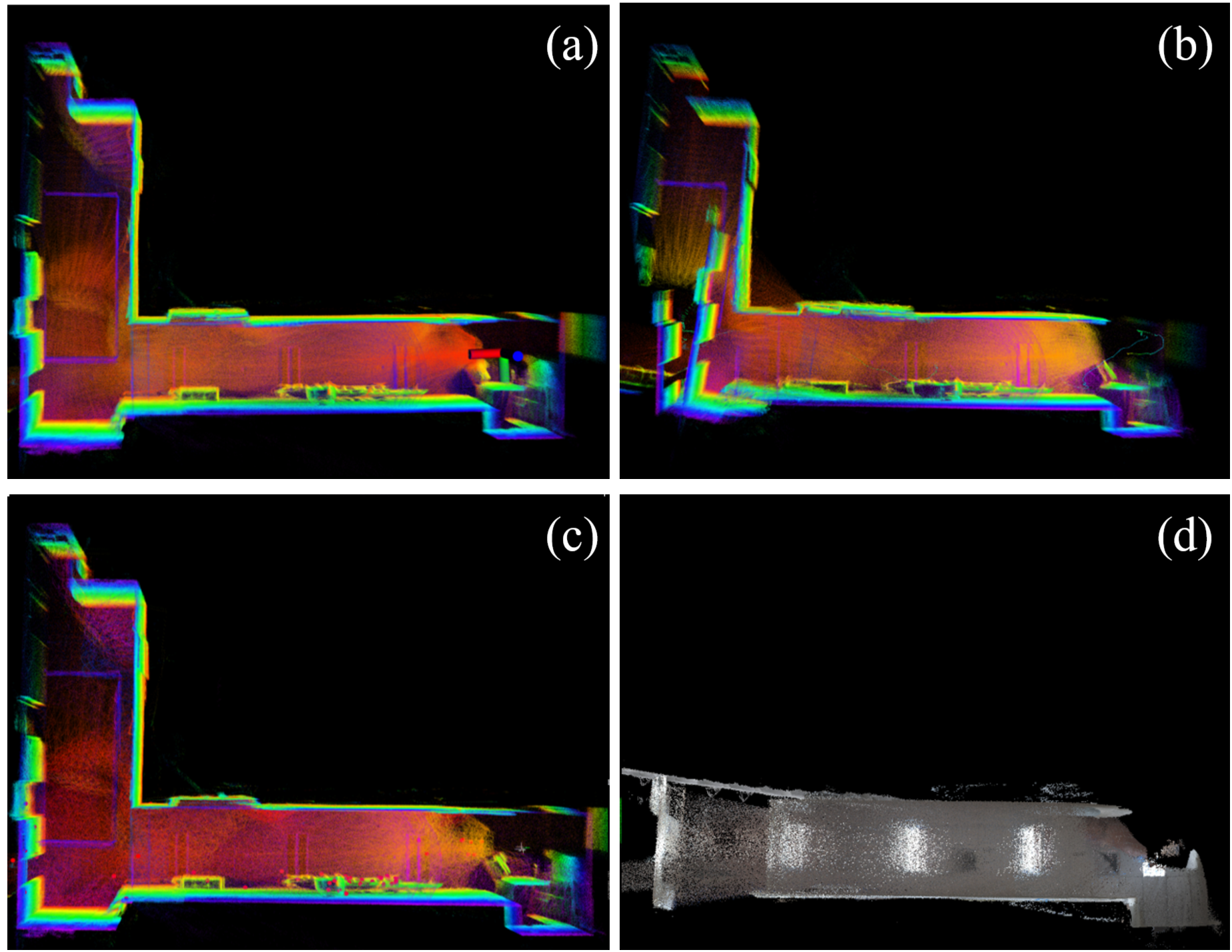}
    \caption{{(a) R3LIVE partially recovers structure with slight misalignment. (b) R2LIVE exhibits pronounced distortion.} (c) F-LIVO preserves the sharp-corner structure. (d) FAST-LIVO2 fails to follow the corner turn when the LiDAR faces the blank wall, resulting in an incomplete map.}
    \label{fig:map}
\end{figure}

{To assess estimator continuity under the specific sensor and geometric-observability degradations considered here, F-LIVO is tested on degraded sequences from the HKU Mars-Lab dataset released with R3LIVE, which include LiDAR scans directed at blank walls, repetitive textures, and camera overexposure. Because complete ground-truth trajectories are unavailable, the approximately coincident initial and final positions are used only to assess accumulated end-to-end drift.}

{The evaluated degeneration sequences primarily represent weak geometric observability rather than a complete absence of planar structures. They include ground-dominated returns, nearly single-directional planar observations, blank-wall conditions, long-range forward motion, and repetitive tunnel geometry. In these cases, planar measurements may still be available, but their spatial and directional diversity is insufficient to constrain all pose degrees of freedom reliably.}



{To independently evaluate the adaptive LiDAR covariance model, we introduce {F-LIVO without adaptive covariance \mbox{(w/o AC)}}. The adaptive and fixed variants use the same base LiDAR residual standard deviation of 0.01. In the fixed variant, this noise value is assigned uniformly to all valid LiDAR point-to-plane residuals, whereas the complete F-LIVO enables the geometry-dependent covariance adjustment. All remaining frontend, association, filter, and map settings are identical, and no sequence-specific parameter tuning is applied.}

{A numerical result is reported whenever a method completes the evaluated interval and produces a valid trajectory, even if the accumulated error is large. The symbol ``--/--'' denotes that the estimator fails to complete the sequence, diverges without a usable trajectory, or does not produce a valid output. A running process alone is not treated as successful completion.} {Therefore, completion status characterizes operational continuity over the evaluated interval rather than general robustness, whereas the reported endpoint error characterizes the accumulated drift of a completed estimate.}

\begin{table}[t]
\centering
\scriptsize
\renewcommand{\arraystretch}{1.15}
\setlength{\tabcolsep}{3.2pt}
\caption{{Accumulated end-to-end translation/rotation drift without loop
closure (cm/deg). F-LIVO (w/o AC) denotes the fixed-noise variant without
adaptive covariance weighting.}}
\label{tab:degeneration_drift}
\begin{tabular}{lccc>{}cc}
\hline
\textbf{Sequence}
& \textbf{FAST-LIVO2}
& \textbf{R3LIVE}
& \textbf{R2LIVE}
& \shortstack{\textbf{F-LIVO}\\\textbf{(w/o AC)}}
& \textbf{F-LIVO} \\
\hline
Degenerate seq00
& 570.0/20.06
& 5.72/0.38
& 23.42/1.43
& 7.113/1.695
& 7.52/0.39 \\

Degenerate seq01
& 1300.6/83.64
& 3.54/0.19
& --/--
& 9.768/1.012
& 4.66/0.25 \\

Degenerate seq02
& --/--
& 8.46/0.28
& --/--
& 21.947/2.251
& 13.68/0.23 \\

LiDAR Degenerate
& --/--
& --/--
& --/--
& 18.7/1.32
& 12.63/0.82 \\

Visual Challenge
& --/--
& --/--
& --/--
& 7.9/0.41
& 6.20/0.26 \\
\hline
\end{tabular}
\end{table}

\begin{figure}[!t]
    \centering
    \includegraphics[width=\linewidth]{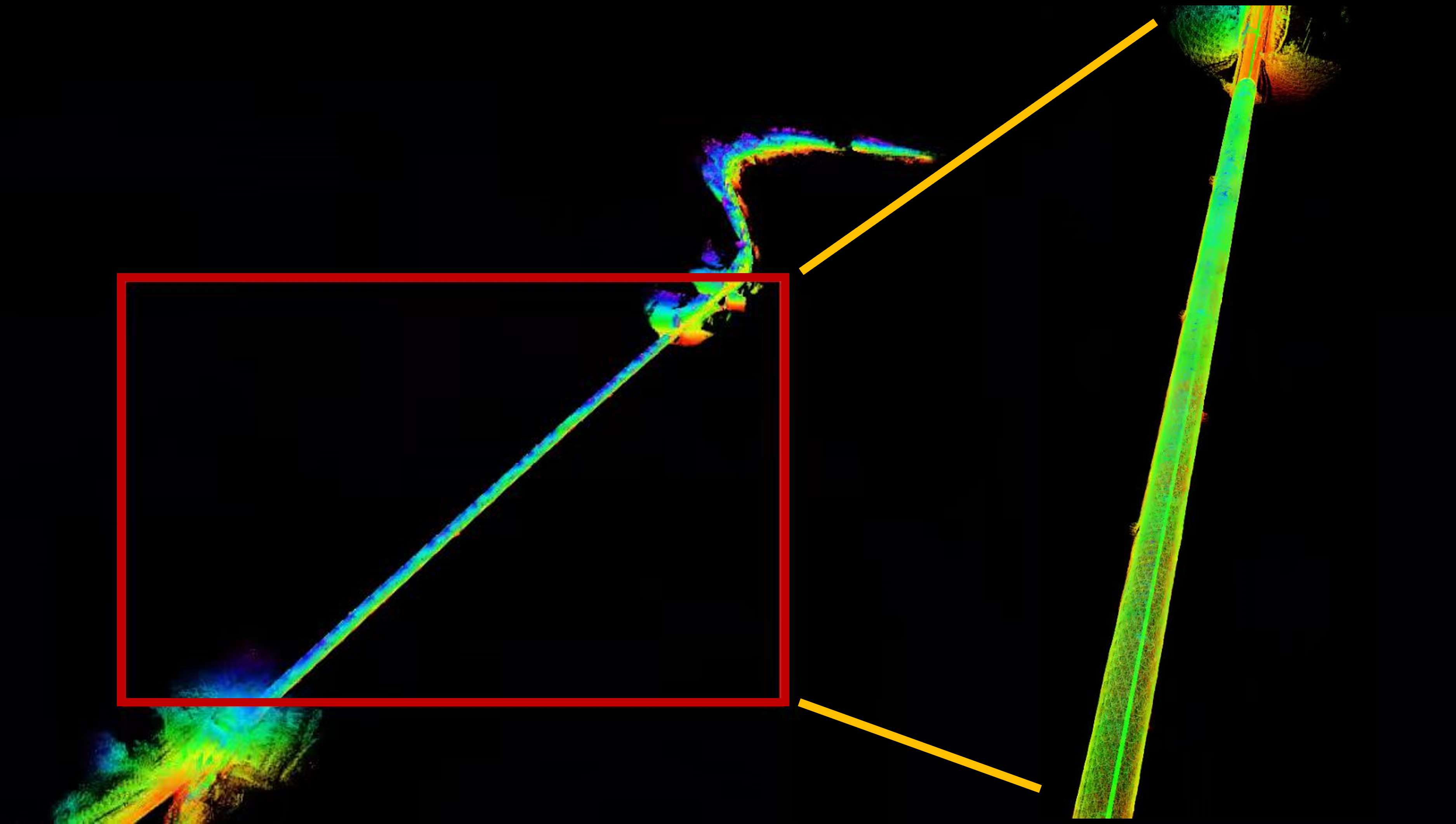}
    \caption{{Qualitative continuity evaluation in a tunnel-like degenerate environment. The red box marks the long, narrow, and repetitive tunnel segment, and the enlarged view shows the corresponding local reconstruction. The pose output and reconstruction remain continuous through the highlighted segment. Since no complete ground-truth trajectory is available, the figure provides qualitative evidence of estimator continuity rather than trajectory-wide accuracy.}}
    \label{fig:tunnel}
\end{figure}

\begin{table}[!t]
\centering
\renewcommand{\arraystretch}{1.25}
\caption{Odometry drift in large-scale environment.}
\label{tab:odometry_drift}
\begin{tabular}{lccc}
\hline
 & \textbf{FAST-LIVO2} & \textbf{F-LIVO} & \textbf{R3LIVE} \\
\hline
\textit{Drift (m/deg)}
 & 2.01~/~0.77
 & 2.03~/~0.74
 & 2.19~/~1.89 \\
\hline
\end{tabular}
\end{table}

\setcounter{table}{5}
{
\begin{table*}[!t]
\centering
\caption{Process-level resource statistics over the common 170-s analysis interval. CPU is aggregate process CPU, where 100\% corresponds to one logical core; memory is process RSS.}
\label{tab:system_occupancy}
\begin{tabular}{lccccc}
\hline
Method & Mean CPU (\%) & Peak CPU (\%) & Mean RSS (GB) & Peak RSS (GB) & Growth rate (MB/min) \\
\hline
F-LIVO & 104.99 & 137.65 & 1.250 & 1.322 & 36.06 \\
FAST-LIVO2 & 243.80 & 332.87 & 3.052 & 5.679 & 1909.48 \\
R3LIVE & 331.52 & 425.49 & 2.611 & 4.742 & 1501.62 \\
R2LIVE & 206.35 & 248.57 & 0.976 & 1.451 & 318.49 \\
\hline
\end{tabular}
\end{table*}
}
\setcounter{table}{4}

\begin{table}[!t]
\centering
\renewcommand{\arraystretch}{1.25}
\caption{MEASUREMENT-UPDATE TIME UNDER DIFFERENT STATE DIMENSIONS (MS).}
\label{tab:runtime_two_methods}
\begin{tabular}{lcccc}
\hline
\textbf{State Size}
 & \textbf{15} & \textbf{30} & \textbf{60} & \textbf{120} \\
\hline
\textit{R3LIVE} & 0.034 & 0.067 & 0.209 & 0.998 \\
\textit{FAST-LIVO2} & 0.023 & 0.024 & 0.026 & 0.028 \\
\textit{F-LIVO} & 0.024 & 0.024 & 0.025 & 0.026 \\
\hline
\end{tabular}
\end{table}
\setcounter{table}{6}

{As shown in Table~\ref{tab:degeneration_drift}, adaptive covariance weighting produces lower accumulated rotation errors on all five evaluated degradation sequences. It also reduces the translation error on Degenerate sequences 01 and 02, the LiDAR-degenerate sequence, and the visual-challenge sequence. On Degenerate sequence 00, the two variants produce comparable translation errors, while the adaptive mode reduces the rotation error from 1.695$^\circ$ to 0.39$^\circ$. On Degenerate sequences 01 and 02, the translation/rotation errors are reduced from 9.768~cm/1.012$^\circ$ to 4.66~cm/0.25$^\circ$ and from 21.947~cm/2.251$^\circ$ to 13.68~cm/0.23$^\circ$, respectively. The same trend is observed in the LiDAR-degenerate and visual-challenge cases. These results indicate that geometry-dependent covariance weighting primarily improves attitude stability by reducing the influence of unreliable local plane observations and can also limit accumulated translational drift under more severe or spatially varying degradation.}

{Together, Tables~\ref{tab:hilt_rmse} and~\ref{tab:degeneration_drift} separately characterize the contributions of the feature-level observation constraints and the adaptive covariance model, respectively.}

{Table~\ref{tab:degeneration_drift} reports accumulated end-to-end translation/rotation drift for three visual--LiDAR degenerate runs, a LiDAR-degenerate run, and a visual-only overexposure challenge. Since the sensor platform returns approximately to its initial pose in these sequences, the drift is computed from the relative transformation between the estimated initial and final poses. No loop closure or global trajectory correction is applied. A small end-to-end drift may still coexist with larger intermediate errors or partial error cancellation along the trajectory.} {Therefore, this metric reports endpoint closure drift rather than trajectory-wide absolute accuracy or the accuracy of intermediate poses. Ground-truth-based trajectory accuracy is evaluated separately using the RMSE/MAX ATE results in Table~\ref{tab:hilt_rmse}; the present metric is used only to characterize endpoint drift, estimator continuity, and failure behavior under the evaluated degradation.}

{The results distinguish estimator continuity from estimation quality. FAST-LIVO2 produces complete trajectories on Degenerate sequences~00 and~01, but the large accumulated rotation errors indicate that these estimates are no longer practically reliable. It fails to complete sequence~02. R2LIVE also diverges in several cases. In contrast, F-LIVO maintains continuous estimation with comparatively low accumulated end-to-end drift across all evaluated degradation cases.}

{Before all evaluations, the camera, LiDAR, and IMU streams are temporally synchronized before being processed by the proposed framework. Residual synchronization errors may still perturb the IMU poses used to reconstruct the visual and LiDAR geometry, thereby affecting nearest-neighbor search, local plane fitting, and cross-modal point-to-plane associations. This sensitivity is more pronounced under rapid rotational or translational motion, whereas the influence is generally smaller under low-dynamic motion and sufficiently accurate synchronization.} R2LIVE's vision front end extracts too few points, resulting in filter divergence. FAST-LIVO2 struggles during the turn-corner maneuver, which leads to a grossly distorted map. {By contrast, F-LIVO's point-to-plane constraints leverage the live 3D scan and planar features extracted in the camera frame to recover depth under LiDAR degradation, thereby maintaining continuous estimation and relatively low accumulated end-to-end drift.}

{In Degenerate sequences~01 and~02, severe image overexposure temporally overlaps with LiDAR observations dominated by ground returns. The overexposed images reduce the number and spatial distribution of reliable visual features, while the ground-dominated point clouds provide insufficient normal-direction diversity for fully constraining the LiDAR update. These sequences therefore represent naturally occurring concurrent visual--LiDAR degradation rather than isolated single-modality failures.}

{Under these concurrent degradation conditions, F-LIVO maintains continuous estimation with accumulated end-to-end translation/rotation drifts of 4.66~cm/0.25$^\circ$ and 13.68~cm/0.23$^\circ$ on sequences~01 and~02, respectively. R2LIVE diverges on both sequences, while FAST-LIVO2 accumulates 1300.6~cm/83.64$^\circ$ on sequence~01 and fails on sequence~02. R3LIVE remains competitive with 3.54~cm/0.19$^\circ$ and 8.46~cm/0.28$^\circ$, respectively.} {These results characterize estimator continuity and the reported accumulated end-to-end drift under the specific concurrent visual and LiDAR observability degradations evaluated here, rather than general robustness or trajectory-wide ground-truth-based accuracy.}

{As an additional qualitative case, Fig.~\ref{fig:tunnel} shows the reconstruction through a long, narrow, and repetitive tunnel segment. The pose output and local reconstruction remain continuous through the highlighted interval without a visually apparent reset or large pose jump. Because no complete ground-truth trajectory is available, this result is interpreted only as qualitative evidence of estimator and reconstruction continuity, rather than as a trajectory-wide accuracy evaluation.}

{F-LIVO constructs a point-to-plane residual only when the local neighborhood passes the geometric validity checks and relies on complementary constraints from accepted LiDAR planes and tracked visual features, together with inertial propagation.} {Its geometric operating boundary is approached when valid planar observations are persistently too sparse to form reliable associations or are concentrated in similar normal directions. The adaptive covariance adjusts the confidence assigned to accepted planar constraints, but it neither restores missing independent constraint directions nor compensates for a prolonged absence of valid planar support. Visual information may still provide complementary constraints; however, when visual tracking is also degraded, the estimate becomes increasingly dominated by inertial propagation, accumulated drift may grow rapidly, and eventual estimator failure can occur.}

{The current plane fitting, cross-modal association, and map update assume that most accepted observations belong to a static environment. Dynamic LiDAR points may bias the fitted local covariance and plane normal, while visual features on moving objects may generate coherent but incorrect cross-modal residuals. Local geometric-validity checks, residual gating, feature rejection, and iterative reassociation can suppress isolated or geometrically inconsistent measurements, but they do not explicitly estimate object motion. The adaptive covariance evaluates local planar reliability rather than motion state; therefore, a moving object may form a well-supported planar surface that passes the planarity checks and remains temporarily consistent with the estimated pose. Repeated acceptance of such observations may corrupt both the state update and the maintained geometric map. Explicit semantic, temporal, or object-motion-based rejection, as investigated by dynamic-scene systems such as Dynam-LVIO~\cite{dynam_lvio_tim}, is outside the scope of the present framework.}

{Overall, F-LIVO maintains continuous estimation with comparatively low accumulated end-to-end drift across the evaluated degradation cases. Its sequence-dependent accuracy and measured process-level resource characteristics provide complementary evidence of practical behavior under the tested conditions; they do not establish a controlled accuracy--resource Pareto trade-off or a component-level memory advantage. Accordingly, the demonstrated robustness should be interpreted only as estimator continuity under the evaluated forms of degraded geometric observability in predominantly static environments where usable local planar constraints remain available, rather than as general robustness to persistent plane scarcity or arbitrary non-planar or dynamic scenes. Prolonged unstructured or plane-scarce scenes in which reliable local planar associations cannot be formed persistently remain outside the demonstrated operating range of the current formulation, especially when visual tracking is also degraded.}

\subsection{Large-scale environment}
In this experiment, we compare F-LIVO with FAST-LIVO2 and R3LIVE on the publicly available R3LIVE campus dataset. The dataset is approximately 1300\,m long and uses ArUco-board-based relative pose references.
{As shown in Tables~\ref{tab:odometry_drift} and~\ref{tab:runtime_two_methods}, F-LIVO maintains accuracy comparable to FAST-LIVO2 in the large-scale outdoor sequence. Table~\ref{tab:runtime_two_methods} further shows that FAST-LIVO2 remains efficient at the measurement-update level, while F-LIVO exhibits the smallest increase in update time and lower update times at the larger state dimensions. R3LIVE shows a substantially more pronounced increase. These results do not imply that measurement-level fusion is inherently inefficient; rather, they show that F-LIVO combines favorable update-time scaling with a simpler persistent cross-modal map representation based on sparse geometric constraints.}

\subsection{Computational Efficiency Experiment}
{We designed a simulation experiment to evaluate how measurement-update time changes with the state dimension when 1000 LiDAR residuals are used. The same residual count and timing protocol are applied to F-LIVO, FAST-LIVO2, and R3LIVE. Table~\ref{tab:runtime_two_methods} is therefore a measurement-update microbenchmark rather than a complete runtime or system-level resource benchmark.}

{As shown in Table~\ref{tab:runtime_two_methods}, FAST-LIVO2 remains efficient over the tested state dimensions, increasing from 0.023~ms to 0.028~ms. F-LIVO exhibits more favorable scaling, increasing only from 0.024~ms to 0.026~ms; it is equal to FAST-LIVO2 at dimension 30 and faster at dimensions 60 and 120. R3LIVE shows a substantially larger increase from 0.034~ms to 0.998~ms. These results confirm that measurement-level fusion is not inherently computationally heavy, while also showing that F-LIVO provides the most favorable measurement-update scaling among the evaluated methods. This microbenchmark is distinct from Fig.~\ref{fig:system_occupancy} and Table~\ref{tab:system_occupancy}, which evaluate complete process-level CPU and RSS.}

\subsection{System occupancy experiment}
{We analyzed process-level CPU usage and resident set size (RSS) over a common 170-s interval after excluding the first 10~s for initialization on
\texttt{hku\_campus\_seq\_00}. As shown in Fig.~\ref{fig:system_occupancy}
and Table~\ref{tab:system_occupancy}, F-LIVO exhibits the lowest mean and peak aggregate CPU usage among the four methods. Compared with FAST-LIVO2, F-LIVO reduces the mean CPU usage from 243.80\% to 104.99\% and the peak CPU usage from 332.87\% to 137.65\%. It also reduces the mean RSS from 3.052~GB to 1.250~GB and the peak RSS from 5.679~GB to 1.322~GB, while decreasing the memory growth rate from 1909.48~MB/min to 36.06~MB/min. Although R2LIVE has a lower mean RSS, its memory increases more rapidly and reaches a higher peak RSS than F-LIVO. These results distinguish CPU demand, resident memory, and memory growth rather than treating them as a single overall footprint measure. Unlike Table~\ref{tab:runtime_two_methods}, which isolates measurement-update scaling, Fig.~\ref{fig:system_occupancy} and Table~\ref{tab:system_occupancy} measure complete process-level CPU and RSS under the evaluated implementations and configurations. {The process-level RSS traces do not by themselves provide an unambiguous decomposition into estimator, map-container, and output-related memory; accordingly, the observed difference is not interpreted as an isolated resource effect of one map component or fusion strategy.} Under the evaluated configuration, F-LIVO achieves the lowest mean and peak CPU usage, the lowest peak RSS, and the lowest RSS growth rate among the evaluated implementations, while R2LIVE retains a lower mean RSS.}

\section{CONCLUSION}
In this paper, we propose a compact filter-based LIVO framework that tightly integrates IMU, camera, and LiDAR observations through feature-level cross-modal constraints. {By leveraging structured geometric features rather than persistent dense photometric associations, F-LIVO achieves competitive, sequence-dependent accuracy and favorable measurement-update scaling. Under the evaluated resource configuration, F-LIVO achieves the lowest mean and peak CPU usage, the lowest peak RSS, and the lowest RSS growth rate among the evaluated implementations, while R2LIVE retains a lower mean RSS. Across the evaluated degradation cases, F-LIVO maintains continuous estimation with comparatively low accumulated end-to-end drift.} {These results do not imply that measurement-level fusion is inherently inefficient, and the process-level resource measurements do not isolate the contribution of the map representation or fusion strategy. The demonstrated robustness is limited to estimator continuity under the evaluated forms of degraded geometric observability in predominantly static environments where usable local planar constraints remain available. Prolonged unstructured or plane-scarce scenes in which reliable local planar associations cannot be formed persistently remain outside the demonstrated operating range, especially when visual tracking is also degraded.}

{The current evaluation is concentrated in predominantly static environments. Explicit handling of coherent moving surfaces and dynamic cross-modal associations remains necessary before the framework can be claimed to operate robustly in highly dynamic urban scenes.}

\bibliographystyle{IEEEtran}


 
\vspace{11pt}


\end{document}